\documentclass{article}

\usepackage[nonatbib, final]{neurips_2022}

\usepackage{url}            % simple URL typesetting

\usepackage{microtype}
\usepackage{graphicx}
\usepackage{subfigure}
\usepackage{booktabs} % for professional tables
\usepackage{amsfonts,amssymb}

\usepackage{hyperref}

\usepackage{amsmath}
\usepackage{amssymb}
\usepackage{mathtools}
\usepackage{amsthm}

\usepackage{algorithm}
\usepackage{algorithmic}

\usepackage[english]{babel}
\usepackage[utf8]{inputenc} % allow utf-8 input
\usepackage[T1]{fontenc}    % use 8-bit T1 fonts
\usepackage{hyperref}       % hyperlinks
\usepackage{amsfonts}       % blackboard math symbols
\usepackage{nicefrac}       % compact symbols for 1/2, etc.
\usepackage{microtype}      % microtypography
\usepackage{xcolor}         % colors

\usepackage{amsmath}
\usepackage{algorithmic}
\usepackage{algorithm}
\usepackage{array}
\usepackage{float}
\usepackage{multirow}
\usepackage{epsfig}
\usepackage{epstopdf}
\usepackage{amssymb}
\usepackage{tabularx}
\usepackage{lineno}
\usepackage{longtable}
\usepackage{bm}
\usepackage{rotating}
\usepackage{mathrsfs}
\usepackage{color}
\usepackage{amsthm}
\usepackage{graphicx}
\usepackage{pythonhighlight}
\usepackage{bm}
\usepackage{stmaryrd}
\usepackage{bbold}
\usepackage{colortbl}
\definecolor{mygray}{gray}{.9}

\usepackage[capitalize,noabbrev]{cleveref}

\theoremstyle{plain}
\newtheorem{theorem}{Theorem}[section]
\newtheorem{proposition}[theorem]{Proposition}

\theoremstyle{definition}

\theoremstyle{remark}

\usepackage[textsize=tiny]{todonotes}

\title{MetaMask: Revisiting Dimensional Confounder for Self-Supervised Learning}

\author{%
  Jiangmeng Li\footnotemark[1] \ , \ Wenwen Qiang\thanks{Equal contributions.} \ , \ Yanan Zhang \\
  University of Chinese Academy of Sciences\\
  Institute of Software Chinese Academy of Sciences\\
  Southern Marine Science and Engineering Guangdong Laboratory (Guangzhou)\\
  \texttt{\{jiangmeng2019, wenwen2018, yanan2018\}@iscas.ac.cn}
  \And
  Wenyi Mo\\
  Gaoling School of Artificial Intelligence\\
  Renmin University of China\\
  \texttt{2022101010@ruc.edu.cn}
  \And
  Changwen Zheng\\
  Institute of Software Chinese Academy of Sciences\\
  Southern Marine Science and Engineering Guangdong Laboratory (Guangzhou)\\
  \texttt{changwen@iscas.ac.cn}
  \And
  Bing Su\thanks{Corresponding author.}\\
  Gaoling School of Artificial Intelligence\\
  Renmin University of China\\
  Beijing Key Laboratory of Big Data Management and Analysis Methods\\
  \texttt{subingats@gmail.com}
  \And
  Hui Xiong\\
  Thrust of Artificial Intelligence\\
  The Hong Kong University of Science and Technology (Guangzhou)\\
  Guangzhou HKUST Fok Ying Tung Research Institute\\
  \texttt{xionghui@ust.hk}
}

\let\oldmaketitle\maketitle
\renewcommand{\maketitle}{\oldmaketitle\setcounter{footnote}{0}}

\begin{document}
\maketitle
\begin{abstract}
As a successful approach to self-supervised learning, contrastive learning aims to learn invariant information shared among distortions of the input sample. While contrastive learning has yielded continuous advancements in sampling strategy and architecture design, it still remains two persistent defects: the interference of task-irrelevant information and sample inefficiency, which are related to the recurring existence of trivial constant solutions. From the perspective of dimensional analysis, we find out that the \textit{dimensional redundancy} and \textit{dimensional confounder} are the intrinsic issues behind the phenomena, and provide experimental evidence to support our viewpoint. We further propose a simple yet effective approach \textit{MetaMask}, short for the \textit{dimensional \textbf{Mask} learned by \textbf{Meta}-learning}, to learn representations against dimensional redundancy and confounder. MetaMask adopts the redundancy-reduction technique to tackle the dimensional redundancy issue and innovatively introduces a dimensional mask to reduce the gradient effects of specific dimensions containing the confounder, which is trained by employing a meta-learning paradigm with the objective of improving the performance of masked representations on a typical self-supervised task. We provide solid theoretical analyses to prove MetaMask can obtain tighter risk bounds for downstream classification compared to typical contrastive methods. Empirically, our method achieves state-of-the-art performance on various benchmarks.\footnote{The implementation is available at \url{https://github.com/jiangmengli/MetaMask}}

% As a successful approach to self-supervised learning, contrastive learning aims to learn invariant information shared among distortions of the input sample. While contrastive learning has yielded continuous advancements in sampling strategy and architecture design, it still remains two persistent defects: the interference of task-irrelevant information and sample inefficiency, which are related to the recurring existence of trivial constant solutions. From the perspective of dimensional analysis, we find out that the dimensional redundancy and dimensional confounder are the intrinsic issues behind the phenomena, and provide experimental evidence to support our viewpoint. We further propose a simple yet effective approach MetaMask, short for the dimensional Mask learned by Meta-learning, to learn representations against dimensional redundancy and confounder. MetaMask adopts the redundancy-reduction technique to tackle the dimensional redundancy issue and innovatively introduces a dimensional mask to reduce the gradient effects of specific dimensions containing the confounder, which is trained by employing a meta-learning paradigm with the objective of improving the performance of masked representations on a typical self-supervised task. We provide solid theoretical analyses to prove MetaMask can obtain tighter risk bounds for downstream classification compared to typical contrastive methods. Empirically, our method achieves state-of-the-art performance on various benchmarks.
\end{abstract}

\section{Introduction} \label{sec:intro}

\begin{figure}
	\centering
	\includegraphics[width=0.95\textwidth]{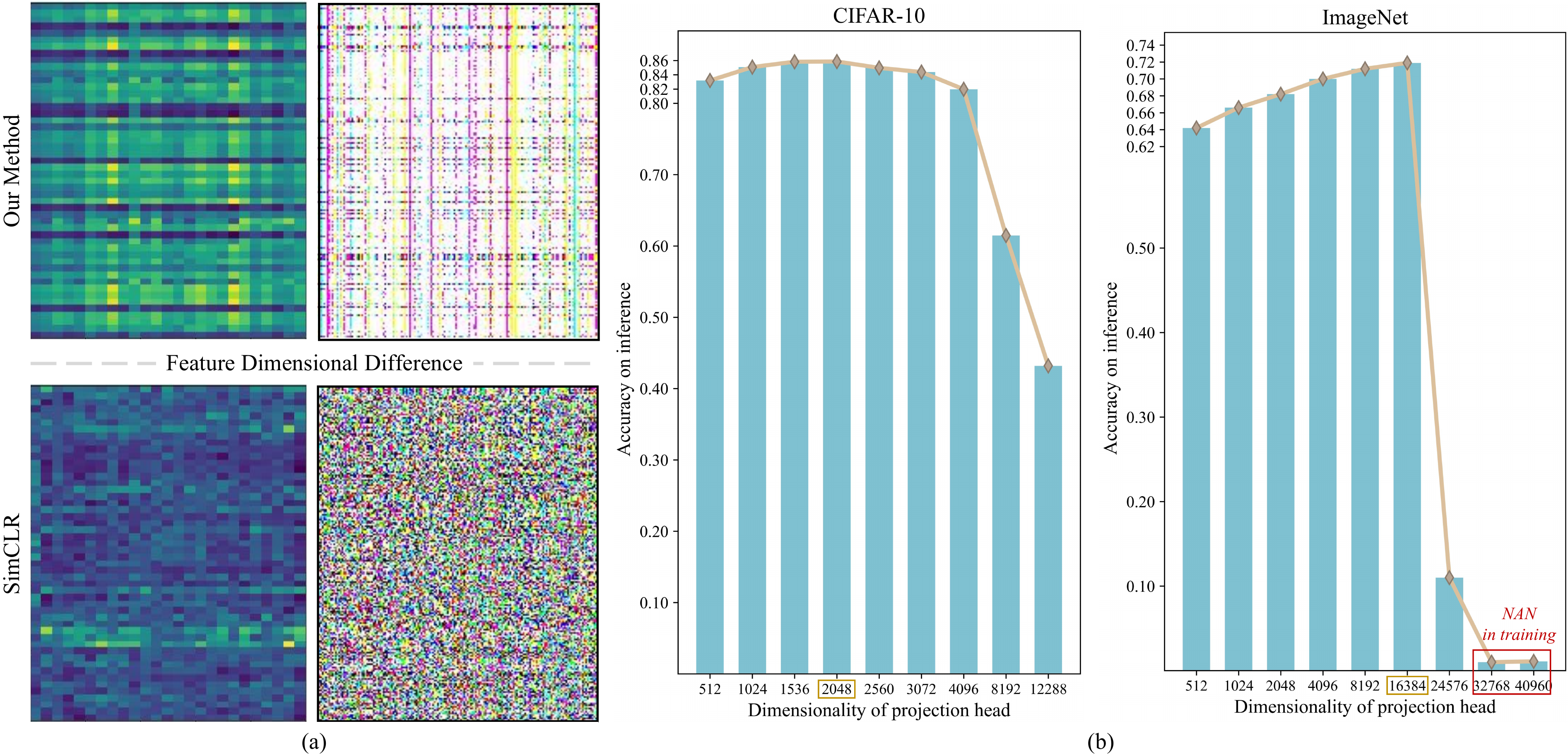}
	\vskip -0.1in
	\caption{(a) The visualization of the representations learned by SimCLR \cite{chen2020simple} and our method using the redundancy-reduction technique \cite{2021Barlow} on CIFAR-10, respectively. The learned features are projected into a color image in RGB format, where different colors represent different types of features. The abscissa axis represents the feature dimensions, and the ordinate axis represents samples of different classes. The greater the color contrast, the lower the dimensional feature similarity. The left plots present the contribution of different dimensions to a specific category classification, and the right plots present the similarity between dimension features within a batch. Compared with SimCLR, our method using redundancy-reduction can indeed learn representations with decoupled dimensions. (b) The experimental results obtained by Barlow Twins on CIFAR-10 and ImageNet datasets. We intuitively increase the dimensionality of the projection head, which leads the dimensions of representations to model more different information. Note that we adopt the official code of Barlow Twins and train on 8 GPUs of NVIDIA Tesla V100.}
	\label{fig:motivdimincrea}
	\vspace{-0.45cm}
\end{figure}

A fundamental idea behind self-supervised learning is to learn discriminative representations from the input data without relying on human annotations. Recent advances in visual self-supervised learning \cite{hjelm2018learning, 2019Philip, 2018RepresentationOord, 2020Mathilde, coconet2022jml} demonstrate that unsupervised approaches can achieve competitive performance over supervised approaches by introducing sophisticated self-supervised tasks. A representative learning paradigm is contrastive learning \cite{Tian2019Contrastive, chen2020simple, 2020Kaiming, 2020Debiased, 2020Hard, metaug2022jml, iclmsr2022jml}, which aims to learn invariant information from different views (generated by data augmentations) by performing instance-level contrast, i.e., pulling views of the same sample together while pushing views of different samples away. Various\cite{2020Bootstrap, 2020WhiteningErmolov, 2021simsiam, mega2022jml} techniques have been explored to address the problem of trivial solutions to self-supervised contrastive learning, e.g., constant representations. However, state-of-the-art methods still suffer from two crucial defects, including the interference of task-irrelevant information and sample inefficiency. The trivial solution problem is related to such defects, e.g., a small number of negative samples or a large proportion of task-irrelevant information may lead the model to learn constant representations. We revisit the current self-supervised learning paradigm from the perspective of dimensional analysis and argue that the intrinsic issues behind the defects of self-supervised learning are the \textit{dimensional redundancy} and \textit{dimensional confounder} \footnote{Throughout this paper, the term \textit{confounder} is used in its idiomatic sense rather than the specific statistical sense in Structural Causal Models \cite{2016Pearl}.}. %To elaborate on our viewpoint, we clarify that each dimension holds a subset of the representation's information entropy. The dimensional redundancy denotes that multiple dimensions hold overlapping information entropy, and the dimensional confounder means that specific dimensions hold task-irrelevant information.

From the perspective of information theory, each dimension holds a subset of the representation's information entropy. The dimensional redundancy denotes that multiple dimensions hold overlapping information entropy. To tackle this issue, Barlow Twins \cite{2021Barlow} takes advantage of a neuroscience approach, i.e., redundancy-reduction \cite{1961barlow}, in self-supervised learning. It makes the cross-correlation matrix of twin embeddings close to the identity matrix, which is conceptually simple yet effective to address the dimensional redundancy issue. Figure \ref{fig:motivdimincrea} (a) supports the superiority of redundancy-reduction in addressing the dimensional redundancy problem. Since Barlow Twins encourages the dimensions of learned representations to model decoupled information, it naturally avoids the collapsed trivial solution that outputs the same vector for all images.

%We revisit the trivial solutions to contrastive learning: the essence of the trivial solution is that the contrastive loss admits collapsed solutions, e.g., outputting the same vector for all images. Yet, Barlow Twins encourages the dimensions of learned representation to model decoupled information, such that this learning paradigm naturally avoids learning collapsed representations.

However, only reducing dimensional redundancy still suffers from the problem of dimensional confounder. The dimensional confounder indicates a set of dimensions that contains ``harmful'' information and further degenerates the performance of the model, e.g., the dimensions simply capturing chaotic and worthless background information from the input. To prove the existence of such dimensional confounders, we conduct motivating experiments from two perspectives. It is stated in \cite{2021Barlow} that Barlow Twins keeps improving as the dimensionality of the projection head increases, since the backbone network acts as a dimensionality bottleneck to constrain the dimensionality of the representation so that the learning paradigm of Barlow Twins can promote the representation capture more task-relevant information with the same dimensionality. Counterintuitively, our observation from the motivating experiments is in stark contrast with the conclusion in \cite{2021Barlow}. As shown in Figure \ref{fig:motivdimincrea} (b), the performance of Barlow twins as a function of the dimensionality presents a convex pattern on either CIFAR-10 or ImageNet. After reaching the peak, the performance gradually decreases as the dimensionality increases until it collapses. We reckon that since the task-relevant semantic information is limited, as the information captured in different dimensions is gradually decoupled, more task-irrelevant noisy information is encoded into the representation, i.e., the dimensional confounder increases, so the performance of the learned representation is getting worse until collapse.

\begin{figure}[t]
	\centering
	\includegraphics[width=0.472\textwidth]{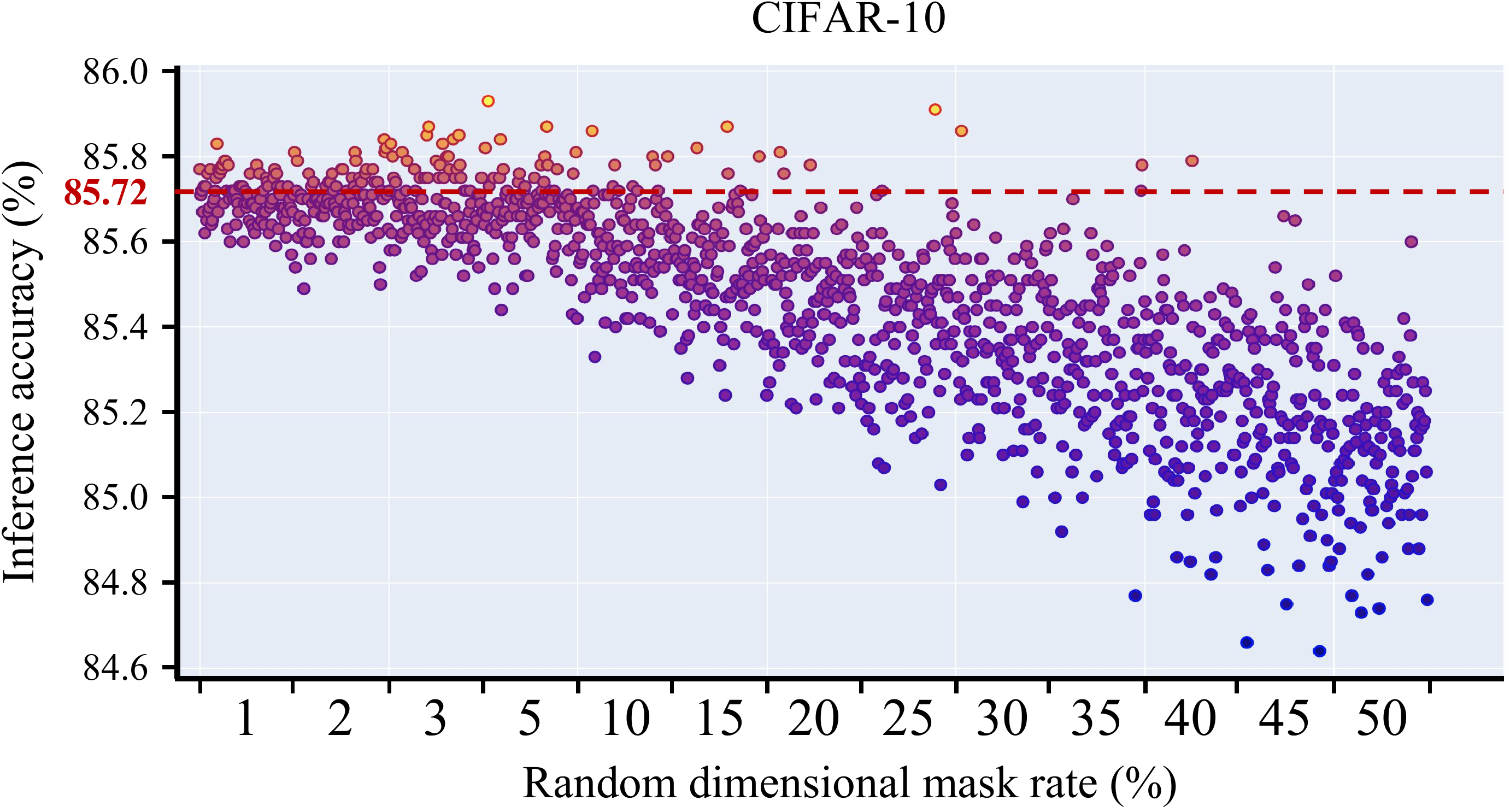}\,\includegraphics[width=0.5225\textwidth]{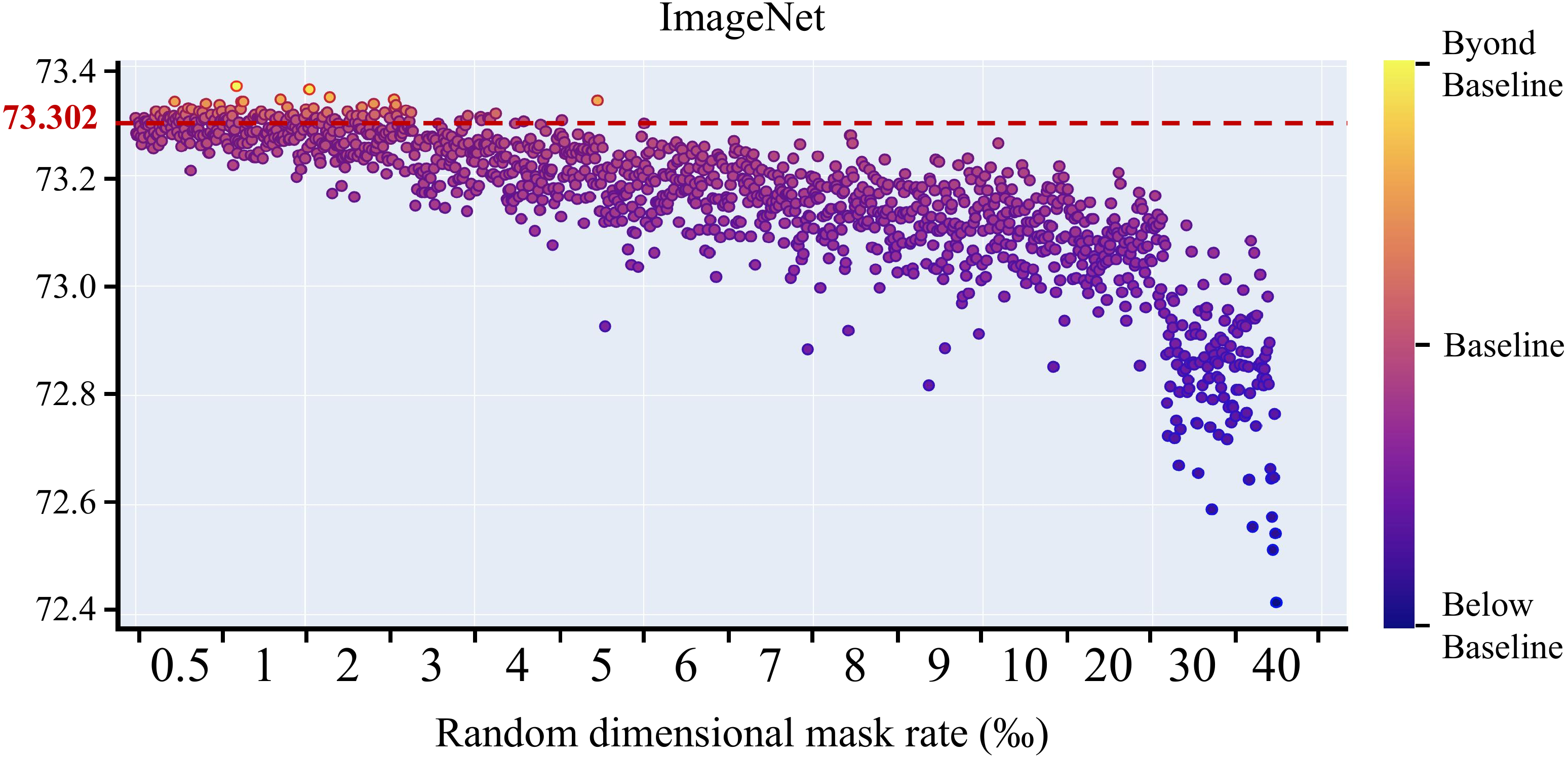}
	\vskip -0.1in
	\caption{Experimental scatter diagrams obtained by Barlow Twins \cite{2021Barlow} with randomly masked dimensions on CIFAR-10 and ImageNet datasets, where \textit{Baseline} and the \textit{\textcolor[RGB]{205,71,119}{red} dashed lines} denote the performance achieved by the unmasked representation of Barlow Twins. Every single point denotes an independent experimental result achieved by randomly masking the original representation at the dimensional level with a specific mask rate. Note that the original representation is consistently fixed. Following the experimental protocol of \cite{2021Barlow, 2021simsiam}, we evaluate the performance on CIFAR-10 and ImageNet by adopting KNN prediction and linear probing, respectively, which further proves the general existence of dimensional confounders.}
	\label{fig:motivmask}
	\vspace{-0.5cm}
\end{figure}

To further validate our inference, we randomly mask several dimensions by frankly setting the masked values to $0$, and then we evaluate the performance of the masked representation. The experimental results are demonstrated in Figure \ref{fig:motivmask}, where every single point denotes the result of a sampled mask. The representations with some specific masked dimensions can indeed achieve better performance than the unmasked representation, and such an observation is consistent on a small-scaled dataset, e.g., CIFAR-10, and a large-scale dataset, e.g., ImageNet. The dimensional mask rate that can improve the performance of the model on ImageNet is on average lower than that on CIFAR-10. We understand that this phenomenon depends on an ingredient: the representation learned on CIFAR-10 models discriminative information for $10$ categories in $512$ dimensions, while the representation uses $2048$ dimensions to model information for $1000$ categories on ImageNet, so the dimension-to-class ratio on CIFAR-10 ($51.2$) is much higher than that on ImageNet ($2.048$). Therefore, the presence of dimensional confounders naturally remains at a low level on ImageNet since supporting $1000$-category classification requires a large amount of heterogeneous discriminative information. This empirical evidence further supports our assumption of the dimensional confounder.
%Another nonnegligible flaw is that even though such an approach is improved with the increase of dimensionality, it 

To this end, we intuitively propose a simple yet effective approach \textit{MetaMask}, short for the \textit{\textbf{Meta}-learned dimensional \textbf{Mask}}, to adjust the impacts of different dimensions in a self-paced manner. In order to learn representations against dimensional redundancy and confounder, we regard the cross-correlation matrix constraint of Barlow Twins as a dimensional regularization term and further introduce a dimensional mask, which simultaneously enhances the gradient effect of the dimensions containing discriminative information and reduces that of the dimensions containing the confounder during training. We employ the meta-learning paradigm to train the learnable dimensional mask with the optimization objective of improving the performance of masked representations on typical self-supervised tasks, e.g., contrastive learning. Consequently, the dimensional mask incessantly adjusts the weights of different dimensions with respect to their gradient contribution to the optimization, so as to make the model focus on the acquisition of discriminative information, which can address the trivial solution caused by treating all dimensions equally in optimization, especially for high-dimensional representations. Empirically, we conduct head-to-head comparisons on various benchmark datasets, which prove the effectiveness of MetaMask. Concretely, the \textbf{contributions} of this paper are four-fold:

\begin{itemize}
	\item We present a timely study on the adaptation of the state-of-the-art self-supervised models on downstream applications and identify two critical issues associated with the learned representation, i.e., dimensional redundancy and dimensional confounder.
	\item We propose a novel approach to tune the gradient effects of dimensions in a self-paced manner during optimization according to their improvements on a self-supervised task.
	\item Theoretically, we provide theorems and proofs to support that MetaMask achieves relatively tighter risk bounds for downstream classification compared to typical contrastive methods.
	\item We conduct extensive experiments to demonstrate the superiority of MetaMask over benchmark methods on downstream tasks.
\end{itemize}

\section{Related Work} \label{sec:relatedwork}
\textbf{Self-supervised learning} achieved impressive success in the field of representation learning without human annotation by constructing auxiliary self-supervised tasks. Typically, self-supervised approaches can be categorized as either generative or discriminative \cite{DBLP:conf/iccv/DoerschGE15, chen2020simple, 2020Bootstrap}. Conventional generative approaches bridge the input data and latent embedding by building a distribution over them, and the learned embeddings are treated as image representations \cite{DBLP:conf/icml/VincentLBM08, DBLP:journals/corr/KingmaW13, DBLP:conf/nips/GoodfellowPMXWOCB14, DBLP:conf/nips/DonahueS19, DBLP:conf/iclr/DumoulinBPLAMC17, DBLP:conf/iclr/BrockDS19}. However, such a learning paradigm learns the distribution at the pixel-level, which is computationally expensive. The detailed information required for image generation may be redundant for representation learning.

For discriminative methods, contrastive learning achieves state-of-the-art performance on various downstream tasks \cite{2020Mathilde, DBLP:conf/nips/Tian0PKSI20, OJ2019Data, DBLP:conf/cvpr/MisraM20, DBLP:conf/iclr/0001ZXH21, DBLP:conf/nips/ChenKSNH20, 2021Vikas, 2021Tete, 2020WhiteningErmolov}. Deep InfoMax \cite{hjelm2018learning} explores to maximize the mutual information between the input and output of an encoder. CPC \cite{2018RepresentationOord} adopts noise-contrastive estimation \cite{2010Michael} as the contrastive loss to train the model by maximizing the mutual information of multiple views, which is deduced by using the Kullback-Leibler divergence \cite{2003Goldbberger}. CMC \cite{Tian2019Contrastive} and AMDIM \cite{2019Philip} employs contrastive learning on the multi-view data. SimCLR \cite{chen2020simple} and MoCo \cite{2020Kaiming, DBLP:journals/corr/abs-2003-04297} use large batch or memory bank to enlarge the amount of available negative features. To sample informative negatives, DebiasedCL\cite{2020Debiased} and HardCL \cite{2020Hard} design sampling strategies by using positive-unlabeled learning methods \cite{2008Charles, 2014Marthinus}. Motivated by \cite{2008Sridharan}, \cite{2020Tsai} provides an information theoretical framework for self-supervised learning and proposes an information bottleneck-based approach. However, these methods rely on a large number of negative samples to learn valuable representations while avoiding trivial solutions, i.e., constant representation. SimSiam \cite{2021simsiam} and BYOL \cite{2020Bootstrap} jointly waive negatives and avoid training degradation by using the techniques of asymmetric siamese network and stop-gradient. SSL-HSIC \cite{DBLP:journals/corr/abs-2106-08320} provides a unified framework for negative-free contrastive learning based on Hilbert Schmidt Independence Criterion \cite{2005HSIC}. The representation learning paradigm employed by current contrastive methods is contrary to the coding criteria proposed by \cite{1961barlow} and \cite{2020yima}, respectively. Therefore, Barlow Twins \cite{2021Barlow} proposes to constrain the cross-correlation matrix between the features to be close to the identity matrix, which can effectively reduce the dimensional redundancy. Yet, such an approach still suffers from the dimensional confounder issue, and we thus propose a new meta-learning-based method to attenuate the over-learning of dimensional confounders by learning a mask to filter out irrelevant information.
%and thereby focus on acquiring discriminative information.

\section{Preliminaries} \label{sec:preliminaries}
Formally, we suppose the input multi-view dataset as $\mathcal{X} = \left\{ \boldsymbol{x}_i^j \Big| i \in \llbracket {1, N^{\ast}} \rrbracket, j \in \llbracket {1, M} \rrbracket \right\}$, where $N^{\ast}$ denotes the number of samples, and $M$ denotes the number of views. Therefore, $\boldsymbol{x}_i^j$ represents the $j$-th view of the $i$-th sample following $\boldsymbol{x}_i^j \thicksim \mathcal{X}$. The multiple views are built by performing identically distributed data augmentations \cite{Tian2019Contrastive, chen2020simple, 2021Barlow}, i.e., $\{\boldsymbol{t}_j\}_{j=1}^M \thicksim \mathcal{T}$.

\subsection{Contrastive Learning}
We recap the preliminaries of contrastive learning \cite{chen2020simple}, which aims to learn an embedding that jointly maximizes agreement between views of the same sample and separates the views of different samples in the latent space. Given a minibatch of $N$ examples sampled from the multi-view dataset $\mathcal{X}$, \textit{positives} represent the pairs $\{\boldsymbol{x}_i^j, \boldsymbol{x}_i^{j^\prime}\}$ where $j, j^\prime \in \llbracket 1, M\rrbracket$, and \textit{negatives} represent $\{\boldsymbol{x}_i^j, \boldsymbol{x}_{i^\prime}^{j^\prime}\}$ where $i \neq i^\prime$. Conventional contrastive methods feed the input $\boldsymbol{x}_i^j$ into a view-shared encoder $f_{\theta}(\cdot)$ to learn the representation $\boldsymbol{h}_i^j$, which is mapped into a feature $\boldsymbol{z}_i^j$ by a projection head $g_{\vartheta}(\cdot)$, where $\theta$ and $\vartheta$ are the network parameters. $f_{\theta}(\cdot)$ and $g_{\vartheta}(\cdot)$ are trained by using the contrastive loss \cite{2018RepresentationOord, chen2020simple}:
\begin{equation}
	{\mathcal{L}_{contrast}} = - \sum_{i \in \llbracket 1, N\rrbracket} \mathop{\sum_{j \in \llbracket 1, M - 1\rrbracket}}\limits_{j^+ \in \llbracket j + 1, M\rrbracket} \log \frac{\exp\left[d\left(\boldsymbol{z}_i^j, \boldsymbol{z}_i^{j^+} \right) / \tau\right]}{\sum_{i^\prime \in \llbracket 1, N \rrbracket} \sum_{j^\prime \in \llbracket 1, M\rrbracket} {\mathbb{1}_{\left[ i \neq i^\prime \lor j \neq j^\prime \right]} \cdot \exp\left[d\left(\boldsymbol{z}_i^j, \boldsymbol{z}_{i^\prime}^{j^\prime} \right) / \tau\right]}}
	\label{eq:cl}
\end{equation}
where $d(\cdot)$ is a discriminating function to measure the similarity of the feature pairs, $\mathbb{1}_{\left[ i \neq i^\prime \lor j \neq j^\prime \right]}$ denotes an indicator function equalling to $1$ if $i \neq i^\prime$ or $j \neq j^\prime$, and $\tau$ is a temperature parameter valued by following \cite{chen2020simple}. In inference, the projection head $g_{\vartheta}(\cdot)$ is discarded, and the representation $\boldsymbol{h}_i^j$ is directly used for downstream tasks.

\begin{figure}
	\centering
	\includegraphics[width=0.95\textwidth]{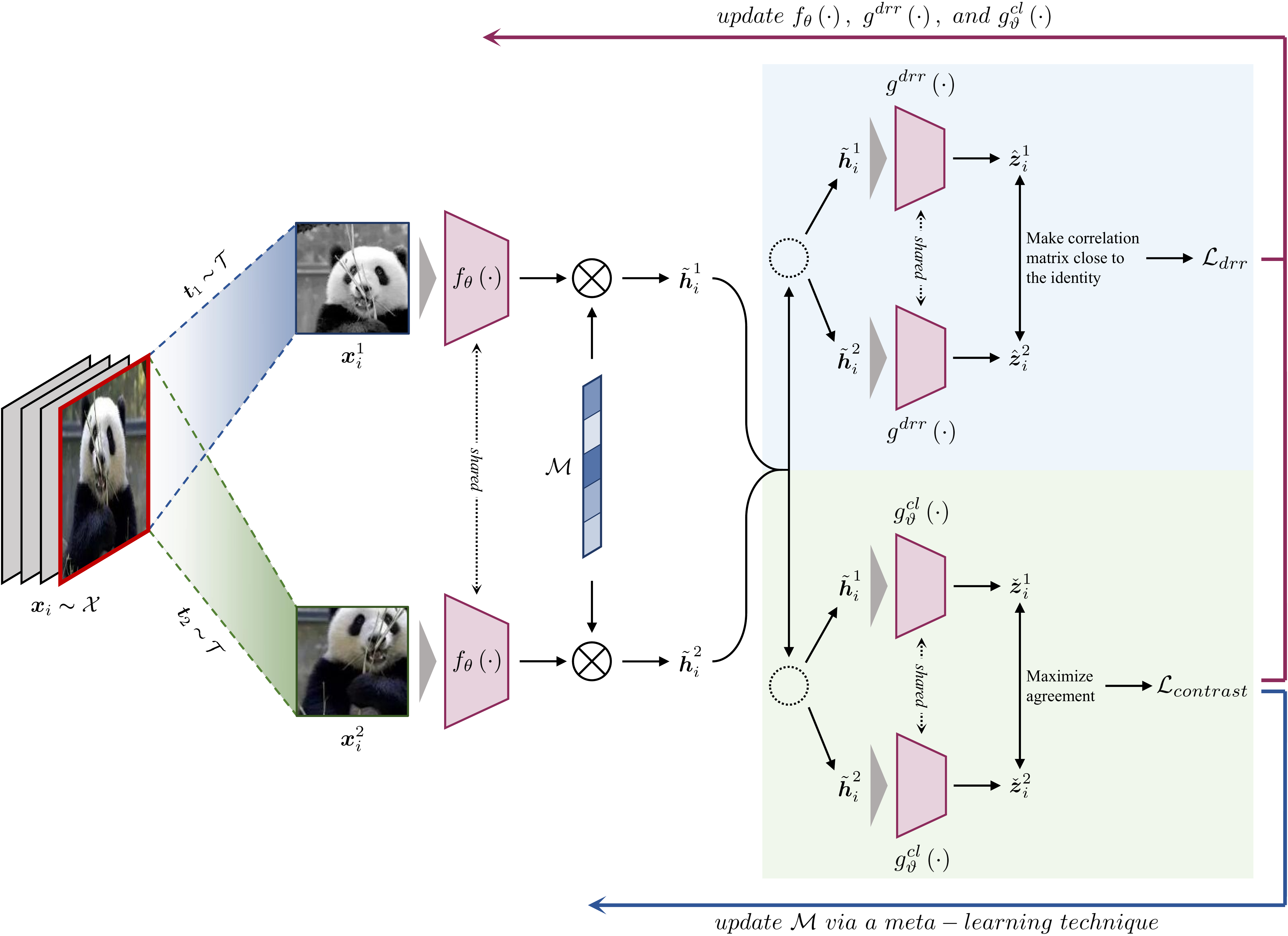}
	\vskip -0.1in
	\caption{The architecture of MetaMask. The solid \textcolor[RGB]{141, 43, 92}{red} line pointing backwards represents the regular training step, where the encoder and projection heads are trained by jointly back-propagating $\mathcal{L}_{drr}$ and $\mathcal{L}_{contrast}$. The solid \textcolor[RGB]{47, 85, 151}{blue} line pointing backwards represents the meta-learning step, where the dimensional mask $\mathcal{M}$ is trained by performing the second-derivative technique on $\mathcal{L}_{contrast}$. All learnable networks are trained until convergence.}
	\label{fig:algoframe}
	\vspace{-0.5cm}
\end{figure}

\subsection{Dimensional Redundant Reduction}
Following the principle proposed by \cite{1961barlow}, \cite{2021Barlow} introduces a redundancy-reduction objective function:
\begin{equation}
	{\mathcal{L}_{drr}} = \mathop{\sum_{j \in \llbracket 1, M - 1\rrbracket}}\limits_{j^+ \in \llbracket j + 1, M\rrbracket} \left[ \sum_{k \in \llbracket 1, D\rrbracket} \left( 1 - \boldsymbol{C}_{kk}^{j, j^+}\right)^2 + \lambda \mathop{\sum_{k, k^\prime \in \llbracket 1, D\rrbracket}}\limits_{k \neq k^\prime} \left( \boldsymbol{C}_{kk^{\prime}}^{j, j^+} \right)^2\right]
	\label{eq:drr}
\end{equation}
where $\lambda$ is a positive hyper-parameter valued by following \cite{2021Barlow}, and $\boldsymbol{C}$ is the cross-correlation matrix computed between the outputs of the encoder along the batch dimension:
\begin{equation}
	\boldsymbol{C}_{kk^\prime}^{j, j^+} = \frac{\sum_{i \in \llbracket 1, N\rrbracket} \boldsymbol{z}_{i,k}^j \cdot \boldsymbol{z}_{i,k^\prime}^{j^+}}{\sqrt{\sum_{i \in \llbracket 1, N\rrbracket} \left( \boldsymbol{z}_{i,k}^j \right)^2} \cdot \sqrt{\sum_{i \in \llbracket 1, N\rrbracket} \left( \boldsymbol{z}_{i,k^\prime}^{j^+}\right)^2}}
	\label{eq:crossmatrix}
\end{equation}
where $k$ and $k^\prime$ index the dimension of features. $\boldsymbol{C}$ is a square matrix with the size of $D$, which denotes the dimensionality of the projected feature $\boldsymbol{z}$. The value range of the elements in $\boldsymbol{C}$ is $\left[-1, 1\right]$, i.e., complete anti-correlation to complete correlation.

\section{Methodology} \label{sec:method}
Our goal is to learn discriminative representations against dimensional redundancy and dimensional confounder from multiple views by performing self-supervised learning. To this end, we propose MetaMask to automatically determine the optimal dimensionality of the representation by decoupling information into different dimensions and filtering out those dimensions with irrelevant information. MetaMask addresses the dimensional redundancy issue by adopting a redundancy-reduction regularization. To tackle the dimensional confounder problem, MetaMask learns a dimensional mask to adjust the gradient weights of dimensions according to their contributions to improving a specific self-supervised task during optimization, which is achieved in a meta-learning manner.

MetaMask's architecture is shown in Figure \ref{fig:algoframe}. Specifically, the sample $\boldsymbol{x}_i^j$ is first fed into the encoder to obtain its representation $\boldsymbol{h}_i^j$. Then, we build a learnable dimensional mask $\mathcal{M} = \left\{ \boldsymbol{\omega}_k \big| k \in \llbracket {1, D} \rrbracket \right\}$, which is introduced to assign a \textit{weight} to each dimension of the representation $\boldsymbol{h}_i^j$ by
\begin{equation}
	\tilde{\boldsymbol{h}}_i^j = \boldsymbol{h}_i^j \otimes \mathcal{M}
	\label{eq:maskh}
\end{equation}
where $\tilde{\boldsymbol{h}}_i^j$ denotes the masked representation, and $\cdot \otimes \cdot$ is an element-wise Hadamard product function. $\tilde{\boldsymbol{h}}_i^j$ is then fed into two separate projection heads $g^{drr}(\cdot)$ and $g^{cl}_{\vartheta}(\cdot)$, respectively. $g^{cl}_{\vartheta}(\cdot)$ plays the same role as in conventional contrastive learning methods, i.e., the contrastive loss $\mathcal{L}_{contrast}$ in Eq. (\ref{eq:cl}) is applied to the output features of $g^{cl}_{\vartheta}(\cdot)$. For the projection head $g^{drr}(\cdot)$, the redundancy-reduction loss $\mathcal{L}_{drr}$ in Eq. (\ref{eq:drr}) is performed to its output features to tackle the dimensional redundancy. 

In training, the encoder $f_{\theta}(\cdot)$, the projection heads $g^{drr}(\cdot)$ and $g^{cl}_{\vartheta}(\cdot)$, as well as the dimensional mask $\mathcal{M}$ are trained in a meta-learning manner consisting of two steps. In the first regular training step, we train $f_{\theta}(\cdot)$, $g^{drr}(\cdot)$ and $g^{cl}_{\vartheta}(\cdot)$ by jointly minimizing the redundancy-reduction and contrastive losses, which is formalized by
\begin{equation}
	\mathcal{L}_{regular} = \mathcal{L}_{drr} + \alpha \cdot \mathcal{L}_{contrast}
	\label{eq:regular}
\end{equation}
where $\alpha$ is a coefficient that controls the balance between $\mathcal{L}_{drr}$ and $\mathcal{L}_{contrast}$, and we evaluate the influence of this hyper-parameter in Section \ref{sec:experiments}.

%On the regular training step, we train the encoder and projection heads by adopting the dimensional redundancy reduction loss in Equation \ref{eq:drr} and the contrastive loss in Equation \ref{eq:cl}. 

In the second meta-learning-based step, we update $\mathcal{M}$ by using the second-derivative technique \cite{2019SelfLiu} to solve a bi-level optimization problem. We encourage $\mathcal{M}$ to \textit{mask} (reduce the gradient weights of) the specific dimensions containing task-irrelevant information, which are regarded as dimensional confounders, so that the encoders can better explore the discriminative information in training. To this end, $\mathcal{M}$ can be updated by computing its gradients with respect to the performance of $f_{\theta}(\cdot)$ and $g^{cl}_{\vartheta}(\cdot)$. The corresponding performance is measured by using the gradients of $f_{\theta}(\cdot)$ and $g^{cl}_{\vartheta}(\cdot)$ during the back-propagation of contrastive loss. Formally, we update the dimensional mask $\mathcal{M}$ by
\begin{equation}
	\mathop{\arg\min}_{\mathcal{M}} \mathcal{L}_{contrast}\left(g^{cl}_{{\vartheta_{trial}}}\left(f_{{\theta_{trial}}}\left(\mathcal{X}^{\prime}\right) \otimes \mathcal{M} \right)\right)
	\label{eq:metam}
\end{equation}
where $\mathcal{X}^\prime$ represents a minibatch sampled from the training dataset $\mathcal{X}$, $\vartheta_{trial}$ and $\theta_{trial}$ represent the \textit{trial} weights of the encoders and projection heads after one gradient update using the contrastive loss defined in Equation \ref{eq:cl}, respectively. We formulate the updating of such trial weights as follows:
\begin{equation}
	\theta_{trial} = \theta - \ell_{\theta}\nabla_{\theta} \mathcal{L}_{contrast}\left(g^{cl}_{{\vartheta}}\left(f_{{\theta}}\left(\mathcal{X}^{\prime}\right) \otimes \mathcal{M} \right)\right), \vartheta_{trial} = \vartheta - \ell_{\vartheta}\nabla_{\vartheta} \mathcal{L}_{contrast}\left(g^{cl}_{{\vartheta}}\left(f_{{\theta}}\left(\mathcal{X}^{\prime}\right) \otimes \mathcal{M} \right)\right)
	\label{eq:fastweight}
\end{equation}
where $\ell_{\theta}$ and $\ell_{\vartheta}$ are learning rates. The intuition behind such a behavior is that we perform a derivative over the derivative (Hessian matrix) of the combination $\{\theta, \vartheta\}$ to update $\mathcal{M}$, i.e., the second-derivative trick. Specifically, we compute the derivative with respect to $\mathcal{M}$ by using a retained computational graph of $\{\theta, \vartheta\}$ and then update $\mathcal{M}$ by back-propagating the derivative as Equation \ref{eq:metam}. 

We train $\mathcal{M}$ only based on the performance of contrastive learning for two reasons: 1) the dimensional redundancy reduction is a regularization term with strong constraints and generally stable convergence, and it is independent of downstream tasks; 2) such a meta-learning-based training approach that promotes contrastive learning performance empowers our method to constrain the upper and lower bounds of the cross-entropy loss on downstream tasks, which ensures that $\mathcal{M}$ can partially mask the gradient contributions of dimensions containing task-irrelevant information and further promote the encoder to focus on learning task-relevant information, which is demonstrated in Section \ref{sec:theory}.

%Following the protocol of meta-learning \cite{2017ModelFinn, 2019SelfLiu},  

%Then, $\mathcal{M}$ is updated by computing its gradients with respect to the performance of $f_{\theta}(\cdot)$ and $g^{cl}_{\vartheta}(\cdot)$. The corresponding performance is measured by using the gradients of $f_{\theta}(\cdot)$ and $g^{cl}_{\vartheta}(\cdot)$ during the back-propagation of contrastive loss. Concretely, all of $f_{\theta}(\cdot)$, $g^{drr}(\cdot)$, $g^{cl}_{\vartheta}(\cdot)$, and $\mathcal{M}$ are iteratively trained until convergence.

The two steps for updating $f_{\theta}(\cdot)$, $g^{drr}(\cdot)$, and $g^{cl}_{\vartheta}(\cdot)$ and updating $\mathcal{M}$ are iteratively repeated until convergence. The training pipeline is detailed by Algorithm \ref{alg:metamask}. 

\begin{algorithm}[t]
	\vskip 0.in
	\begin{algorithmic}
		\STATE {\bfseries Input:} Multi-view dataset $\mathcal{X}$, minibatch size $N$, and a hyper-parameter $\alpha$.\\
		\STATE {\bf Initialize} The neural network parameters: $\theta$ for $f_{\theta}(\cdot)$, $\vartheta$ for $g^{cl}_{\vartheta}(\cdot)$, $\vartheta^{drr}$ for $g^{drr}(\cdot)$, and $\mathcal{M} = \left\{ \boldsymbol{\omega}_k \big| k \in \llbracket {1, D} \rrbracket \right\}$. The learning rates: $\ell_\theta$ and $\ell_\vartheta$, etc.
		\REPEAT
		\FOR{$t$-th training iteration}
		\STATE Iteratively sample a minibatch $\mathcal{X}^\prime$ with $N$ examples from $\mathcal{X}$.
		\STATE $\# \ regular \ training \ step, \ fix \ \mathcal{M}$
		\STATE $\mathop{\arg\min}_{\theta, \vartheta, \vartheta^{drr}} \mathcal{L}_{drr}\left(g^{drr}_{{\vartheta^{drr}}}\left(f_{{\theta}}\left(\mathcal{X}^{\prime}\right) \otimes \mathcal{M} \right)\right) + \alpha \cdot \mathcal{L}_{contrast}\left(g^{cl}_{{\vartheta}}\left(f_{{\theta}}\left(\mathcal{X}^{\prime}\right) \otimes \mathcal{M} \right)\right)$ \\
		%\ENDFOR
		%\FOR{$t$-th training iteration}
		%\STATE Iteratively sample a minibatch $\mathcal{X}^\prime$ with $N$ examples from $\mathcal{X}$.
		\STATE $\# \ compute \ trial \ weights \ and \ retain \ computational \ graph \ , \ fix \ \theta \ and \ \vartheta$
		\STATE $\theta_{trial} = \theta - \ell_{\theta}\nabla_{\theta} \mathcal{L}_{contrast}\left(g^{cl}_{{\vartheta}}\left(f_{{\theta}}\left(\mathcal{X}^{\prime}\right) \otimes \mathcal{M} \right)\right)$ \\
		\STATE $\vartheta_{trial} = \vartheta - \ell_{\vartheta}\nabla_{\vartheta} \mathcal{L}_{contrast}\left(g^{cl}_{{\vartheta}}\left(f_{{\theta}}\left(\mathcal{X}^{\prime}\right) \otimes \mathcal{M} \right)\right)$ \\
		\STATE $\# \ meta \ training \ step \ using \ second \ derivative$
		\STATE $\mathop{\arg\min}_{\mathcal{M}} \mathcal{L}_{contrast}\left(g^{cl}_{{\vartheta_{trial}}}\left(f_{{\theta_{trial}}}\left(\mathcal{X}^{\prime}\right) \otimes \mathcal{M} \right)\right)$ \\
		\ENDFOR
		\UNTIL $\theta$, $\vartheta$, $\vartheta^{drr}$, and $\mathcal{M}$ converge.
	\end{algorithmic}
	\vskip -0.in
	\caption{MetaMask}
	\label{alg:metamask}
\end{algorithm}

\section{Theoretical Analyses} \label{sec:theory}

%\begin{equation}
%	\begin{aligned}
%		&\mathcal{L}_{contrast}\left(g^{cl}_{\vartheta}\left(\boldsymbol{h}\right)\right) - \sqrt{\Phi\left(g^{cl}_{\vartheta}\left(\boldsymbol{h}\right) \big| \boldsymbol{y}\right)} - \frac{1}{2} \cdot \sum_{k \in \llbracket 1, D \rrbracket} \sqrt{\Phi\left(\lfloor g^{cl}_{\vartheta}\left(\boldsymbol{h}\right)\rceil^k \big| \boldsymbol{y}\right)} - \mathcal{O} \left( M^{- \frac{1}{2}} \right) \\ &\leq \mathcal{L}_{crossentropy}\left(\boldsymbol{h}\right) + \log \left(\frac{M}{D}\right) \leq 	\mathcal{L}_{contrast}\left(g^{cl}_{\vartheta}\left(\boldsymbol{h}\right)\right) + \sqrt{\Phi\left(g^{cl}_{\vartheta}\left(\boldsymbol{h}\right) \big| \boldsymbol{y}\right)} + \mathcal{O} \left( M^{- \frac{1}{2}} \right)
%	\end{aligned}
%\end{equation} (, proposed by \cite{wang2022chaos})

Current contrastive methods follow the fundamental assumption of multi-view learning (Assumption 1 in \cite{2008Sridharan}, Assumption 1 in \cite{2020Tsai}, and Assumption 4.1 in \cite{wang2022chaos}), i.e., views (or distortions) of a same sample holds the invariant discriminative label-relevant information. Theorem 4.2 in \cite{wang2022chaos} further proposes the guarantees for general encoders in the learning paradigm of regular contrastive learning, which proves that the contrastive loss in the self-supervised learning stage can constrain the upper and lower bounds of the cross-entropy loss in the supervised learning stage for downstream tasks. To elaborate the behavior of our MetaMask, we extend this Theorem to build a connection between the proposed masked representation and downstream performance by
\begin{theorem}
	(Connecting Masked Representations to Downstream Cross-Entropy Loss). Under the general assumption of self-supervised multi-view learning, when $\mathcal{M}$ is optimal, for any coupled $\boldsymbol{h}, \tilde{\boldsymbol{h}} \in \mathbb{R}$, $\mathcal{L}_{crossentropy}\left({\boldsymbol{h}}\right)$ for downstream classification can be bounded by $\mathcal{L}_{contrast}\left(g^{cl}_{\vartheta}\left(\tilde{\boldsymbol{h}}\right)\right)$ in self-supervised learning:
	\begin{equation}
		\begin{aligned}
			&\mathcal{L}_{contrast}\left(g^{cl}_{\vartheta}\left(\tilde{\boldsymbol{h}}\right)\right) - \sqrt{\Phi\left(g^{cl}_{\vartheta}\left(\tilde{\boldsymbol{h}}\right) \Big| \boldsymbol{y}\right)} - \frac{1}{2} \cdot \sum_{k \in \llbracket 1, D \rrbracket} {\Phi\left(\lfloor g^{cl}_{\vartheta}\left(\tilde{\boldsymbol{h}}\right)\rceil^k \Big| \boldsymbol{y}\right)} - \mathcal{O} \left( M^{- \frac{1}{2}} \right) \\ &\leq \mathcal{L}_{crossentropy}\left({\boldsymbol{h}}\right) + \log \left(\frac{M}{D}\right) \leq 	\mathcal{L}_{contrast}\left(g^{cl}_{\vartheta}\left(\tilde{\boldsymbol{h}}\right)\right) + \sqrt{\Phi\left(g^{cl}_{\vartheta}\left(\tilde{\boldsymbol{h}}\right) \Big| \boldsymbol{y}\right)} + \mathcal{O} \left( M^{- \frac{1}{2}} \right)
		\end{aligned}
	\end{equation}
	where $M$ denotes the quantity of negative samples, $D$ denotes the dimensionality of the representation, $\boldsymbol{h}$ denotes the unmasked representation, $\tilde{\boldsymbol{h}}$ denotes the masked representation obtained by Equation \ref{eq:maskh}, $\boldsymbol{y}$ denotes the \textit{target} label, $\Phi\left(\cdot | \boldsymbol{y} \right) = \mathbb{E}_{\mathcal{P}\left(y\right)}\left[\mathbb{E}_{\mathcal{P}\left( \cdot  | y\right)} \delta \left( \cdot, \mathbb{E}_{\mathcal{P}\left( \cdot  | y\right)} \right) \right]$ represents the conditional variance function where $\delta\left(\cdot, \cdot\right)$ is a discriminating function to measure the difference between the terms, e.g., $\delta\left(\cdot, \cdot\right) = \left\Vert\left[\cdot\right] - \left[\cdot\right]\right\Vert^2$ in low-dimensional embedding space or $\delta\left(\cdot, \cdot\right) = - \log\left(\cos\left(\cdot, \cdot\right)\right) = - \log\frac{\left[\cdot\right] \times \left[\cdot\right]}{||\cdot|| \times ||\cdot||}$ in high-dimensional embedding space, $\lfloor \cdot \rceil^k$ is a function acquiring $k$-th dimension vector, $\log \left(\frac{M}{D}\right)$ is a constant, and $\mathcal{O} \left( M^{- \frac{1}{2}} \right)$ is the approximation error's order.
	\label{the:connect}
\end{theorem}

Theorem \ref{the:connect} states that reducing the risk of contrastive learning can improve the performance on downstream tasks, which further supports our intuition to make the model focus on the acquisition of discriminative information by learning a dimensional mask with the objective of improving the performance of masked representations on contrastive learning is theoretically sound. Refer to Appendix \ref{app:discusion} for the elaboration of MetaMask's behavior to mask the gradient contribution of specific dimensions containing confounder information. Hence, we derive
\begin{theorem}
	(Guarantees for Reduced Conditional Variance of Masked Representations). When $\mathcal{M}$ is optimal, for any coupled $\boldsymbol{h}, \tilde{\boldsymbol{h}} \in \mathbb{R}$, given label $\boldsymbol{y}$, the conditional variance of $\tilde{\boldsymbol{h}}$ is reduced:
	\label{the:tighter}
	\begin{equation}
		\Phi\left(g^{cl}_{\vartheta}\left(\tilde{\boldsymbol{h}}\right) \Big| \boldsymbol{y}\right) \leq \Phi\left(g^{cl}_{\vartheta}\left({\boldsymbol{h}}\right) \Big| \boldsymbol{y}\right), \ yet \ \ \ \Phi\left(\lfloor g^{cl}_{\vartheta}\left(\tilde{\boldsymbol{h}}\right)\rceil^k \Big| \boldsymbol{y}\right) \cong \Phi\left(\lfloor g^{cl}_{\vartheta}\left({\boldsymbol{h}}\right)\rceil^k \Big| \boldsymbol{y}\right).
	\end{equation}
\end{theorem}
Theorem \ref{the:tighter} states that given the label $y$, the masked representation $\tilde{\boldsymbol{h}}$ has smaller conditional variance than $\boldsymbol{h}$ in contrastive learning. We bring Theorem \ref{the:tighter} into Theorem \ref{the:connect} to derive a conclusion: our approach can better bound the downstream classification risk, i.e., the upper and lower bounds of supervised cross-entropy loss obtained by MetaMask are tighter than typical contrastive learning methods. Please refer to Appendix \ref{app:proof} for the corresponding proofs.

\begin{table}[t]
	\tiny
	\renewcommand\arraystretch{1.1}
	\vskip 0.1in
	\caption{Comparison of different methods on classification accuracy (top 1). We follow the experimental setting of \cite{Tian2019Contrastive} and adopt \textit{conv} and \textit{fc} as the backbones in the experiments.}
	\vskip -0.14in
	\label{tab:convfc}
	\setlength{\tabcolsep}{0.5pt}
	\begin{center}
		\begin{small}
			\begin{tabular}{llllllllll}
				\toprule
				\multicolumn{2}{c}{\multirow{2}*{Model}} & \multicolumn{2}{c}{IN-200 \cite{krizhevsky2009learning}} & \multicolumn{2}{c}{STL-10 \cite{coates2011analysis}} & \multicolumn{2}{c}{CIFAR-10 \cite{krizhevsky2009learning}} & \multicolumn{2}{c}{CIFAR-100 \cite{krizhevsky2009learning}} \\ 
				\cline{3-10}
				& & conv  & fc & conv & fc& conv & fc& conv & fc \\
				\hline
				\hline
				\multicolumn{2}{l}{Barlow Twins \cite{2021Barlow}} & 39.81 & 40.34 & 80.97 & 81.43 & 76.63 & 78.49 & 52.80 & 52.95 \\ \rowcolor{mygray}
				\multicolumn{2}{c}{+ MetaMask$_{\textcolor{red}{+4.95}}$} & 42.91$_{\textcolor{red}{+3.10}}$ & 42.06$_{\textcolor{red}{+1.72}}$ & 84.96$_{\textcolor{red}{+3.99}}$ & 86.13$_{\textcolor{red}{+4.70}}$ & 85.24$_{\textcolor{red}{+8.61}}$ & 86.90$_{\textcolor{red}{+8.41}}$ & 58.02$_{\textcolor{red}{+5.22}}$ & 56.78$_{\textcolor{red}{+3.83}}$ \\ 
				\hline
				\hline
				\text{SwAV \cite{2020Mathilde}} & & 39.67 & 39.02 & 74.65 & 75.30 & 72.19 & 72.36 & 46.58 & 46.75 \\ \rowcolor{mygray}
				\multicolumn{2}{c}{+ MetaMask$_{\textcolor{red}{+5.00}}$} & 40.28$_{\textcolor{red}{+0.61}}$ & 40.05$_{\textcolor{red}{+1.03}}$ & 81.34$_{\textcolor{red}{+6.69}}$ & 82.19$_{\textcolor{red}{+6.89}}$ & 80.37$_{\textcolor{red}{+8.18}}$ & 77.22$_{\textcolor{red}{+4.86}}$ & 52.07$_{\textcolor{red}{+5.49}}$ & 52.98$_{\textcolor{red}{+6.23}}$ \\ 
				\text{SimCLR \cite{chen2020simple}} & & 36.24 & 39.83 & 75.57 & 77.15 & 80.58 & 80.07 & 50.03 & 49.82 \\ \rowcolor{mygray}
				\multicolumn{2}{c}{+ MetaMask$_{\textcolor{red}{+3.60}}$} & 40.56$_{\textcolor{red}{+4.32}}$ & 40.82$_{\textcolor{red}{+0.99}}$ & 81.37$_{\textcolor{red}{+5.80}}$ & 82.90$_{\textcolor{red}{+5.75}}$ & 82.85$_{\textcolor{red}{+2.27}}$ & 81.64$_{\textcolor{red}{+1.57}}$ & 53.70$_{\textcolor{red}{+3.67}}$ & 54.24$_{\textcolor{red}{+4.42}}$ \\ 
				\text{CMC \cite{Tian2019Contrastive}} & & 41.58 & 40.11 & 83.03 & 85.06 & 81.31 & 83.28 & 58.13 & 56.72 \\ \rowcolor{mygray}
				\multicolumn{2}{c}{+ MetaMask$_{\textcolor{red}{+1.73}}$} & 42.91$_{\textcolor{red}{+1.33}}$ & 42.06$_{\textcolor{red}{+1.95}}$ & 84.96$_{\textcolor{red}{+1.93}}$ & 86.13$_{\textcolor{red}{+1.07}}$ & 85.24$_{\textcolor{red}{+3.93}}$ & 86.90$_{\textcolor{red}{+3.62}}$ & 58.02$_{\textcolor{blue}{-0.09}}$ & 56.78$_{\textcolor{red}{+0.06}}$ \\ 
				\text{BYOL \cite{2020Bootstrap}} & & 41.59 & 41.90 & 81.73 & 81.57 & 77.18 & 80.01 & 53.64 & 53.78 \\ \rowcolor{mygray}
				\multicolumn{2}{c}{+ MetaMask$_{\textcolor{red}{+1.68}}$} & 42.72$_{\textcolor{red}{+1.13}}$ & 43.29$_{\textcolor{red}{+1.39}}$ & 83.01$_{\textcolor{red}{+1.28}}$ & 83.44$_{\textcolor{red}{+1.87}}$ & 79.26$_{\textcolor{red}{+2.08}}$ & 82.73$_{\textcolor{red}{+2.72}}$ & 55.68$_{\textcolor{red}{+2.04}}$ & 54.72$_{\textcolor{red}{+0.94}}$ \\ 
				\text{SimSiam \cite{2021simsiam}} & & 41.03 & 41.27 & 80.91 & 81.88 & 78.14 & 81.13 & 52.55 & 53.52 \\ \rowcolor{mygray}
				\multicolumn{2}{c}{+ MetaMask$_{\textcolor{red}{+1.44}}$} & 41.82$_{\textcolor{red}{+0.79}}$ & 42.67$_{\textcolor{red}{+1.40}}$ & 82.35$_{\textcolor{red}{+1.44}}$ & 83.96$_{\textcolor{red}{+2.08}}$ & 80.03$_{\textcolor{red}{+1.89}}$ & 81.39$_{\textcolor{red}{+0.26}}$ & 54.70$_{\textcolor{red}{+2.15}}$ & 55.02$_{\textcolor{red}{+1.50}}$ \\ 
				\text{DCL \cite{2020Debiased}} & & 38.79 & 40.26 & 77.09 & 78.39 & 80.89 & 80.93 & 51.38 & 51.09 \\ \rowcolor{mygray}
				\multicolumn{2}{c}{+ MetaMask$_{\textcolor{red}{+1.83}}$} & 40.28$_{\textcolor{red}{+1.49}}$ & 41.69$_{\textcolor{red}{+1.43}}$ & 80.94$_{\textcolor{red}{+3.85}}$ & 81.06$_{\textcolor{red}{+2.67}}$ & 81.33$_{\textcolor{red}{+0.44}}$ & 81.34$_{\textcolor{red}{+0.41}}$ & 53.05$_{\textcolor{red}{+1.67}}$ & 53.74$_{\textcolor{red}{+2.65}}$ \\ 
				\text{HCL \cite{2020Hard}} & & 40.05 & 41.23 & 79.86 & 80.20 & 82.13 & 82.76 & 52.69 & 53.13 \\ \rowcolor{mygray}
				\multicolumn{2}{c}{+ MetaMask$_{\textcolor{red}{+1.14}}$} & 40.43$_{\textcolor{red}{+0.38}}$ & 41.01$_{\textcolor{blue}{-0.22}}$ & 81.52$_{\textcolor{red}{+1.66}}$ & 82.09$_{\textcolor{red}{+1.89}}$ & 83.65$_{\textcolor{red}{+1.52}}$ & 83.95$_{\textcolor{red}{+1.19}}$ & 54.03$_{\textcolor{red}{+1.34}}$ & 54.26$_{\textcolor{red}{+1.13}}$ \\
				\bottomrule
			\end{tabular}
		\end{small}
	\end{center}
	\vskip -0.2in
\end{table}

\section{Experiments} \label{sec:experiments}
%We compare our MetaMask against benchmark methods on the established datasets.

\subsection{Benchmarking MetaMask with Various Backbones}
\textbf{Implementations.} For the comparisons demonstrated in Table \ref{tab:convfc}, we uniformly set the batch size as 64, and we adopt a network with the 5 convolutional layers in AlexNet \cite{2012Krizhevsky} as \textit{conv} and a network with 2 additional fully connected layers as \textit{fc}. We straightforwardly adopt the same data augmentations in \cite{Tian2019Contrastive} and \cite{li2022metaug}, and the memory bank \cite{un2} is adopted to facilitate calculations by retrieving 4,096 past negative features during training. Following the experimental principle of linear probing, we remove the projection head and re-train an MLP as the classification head for the learned representation.

For the experiments in Table \ref{tab:resnet18}, the batch size is valued by 512, and ResNet-18 \cite{He_2017_ICCV} is used as the backbone encoder. We adopt the data augmentation and other experimental settings following \cite{2021Barlow}. Note that the KNN classifier is adopted for evaluation. %the adopted evaluation approach is KNN prediction.

\textbf{Discussion of the comparisons using conv and fc encoders.} Table \ref{tab:convfc} shows the comparisons on four benchmark datasets, and the corresponding positive or negative margins are reported. We observe that MetaMask beats the best prior methods by a significant margin on most downstream tasks, which demonstrates that MetaMask can efficiently improve the performance of the learned representation when the self-supervision is insufficient, i.e., the limited batch size and weak encoder.

\textbf{Discussion of the comparisons using the ResNet encoder.} We further perform classification comparisons using a strong encoder, i.e., ResNet. Table \ref{tab:resnet18} shows that MetaMask can consistently improve the state-of-the-art methods, which indicates that the proposed method has strong adaptability to different encoders. The experimental evidence proves our assumption of the dimensional confounder. Thus, given a fixed dimensionality of the projection head, MetaMask can indeed improve the discriminativeness of the learned representation by \textit{masking} the dimensional confounder.

\subsection{Validation of the Robustness of MetaMask against Dimensional Confounder} \label{sec:valrobust}
We conduct experiments to prove that MetaMask is able to mitigate the negative impact of the dimensional confounder on the discriminativeness of the learned representations, and the results are shown in Figure \ref{fig:robust}. The intriguing outcome demonstrates two merits of the proposed method: 1) even if the dimensionality of the projection head changes, MetaMask is consistently robust to the dimensional confounder, i.e., the model maintains relatively consistent performance; 2) MetaMask can alleviate the negative impact of \textit{curse of dimensionality} \cite{1964DYNAMIC, 2014Vincent}, e.g., given a fixed number of examples, as the dimensionality of the representation increases during training, the performance of the model on downstream tasks instead degenerates. Concretely, Figure \ref{fig:motivmask} demonstrates that regardless of whether representation learning suffers from the curse of dimensionality, dimensionality confounding factors are widely present in learned representations, and hence, the defined dimensional confounder is a more generalized issue of the curse of dimensionality. Refer Appendix \ref{app:diffcurse} for the further discussion. Moreover, we observe a surprising result that MetaMask$^\ast$ achieves better performance. We provide an understandable explanation: the essence of the curse of dimensionality is that a network with a large number of parameters is prone to over-fitting to the training data. Therefore, compared with the annealing strategy, a fixed and relatively large learning rate can prevent the model from falling into a local optimum (over-fitting) during optimization. Our further experiments show that MetaMask$^\ast$ collapses abruptly at larger dimensional settings, e.g., 16384, which demonstrates that the trick of using a fixed learning rate can only unsteadily improve the performance of MetaMask, but this behavior is \textit{not} essential. Figure \ref{fig:robust} shows that under the different experimental settings, vanilla MetaMask consistently counteracts the negative effects of the dimensional confounder.

\begin{table}[t]
	\tiny
	\renewcommand\arraystretch{1.1}
	\vskip 0.in
	\begin{minipage}{\textwidth}
		\begin{minipage}[t]{0.7\textwidth}
			\makeatletter\def\@captype{table}
			\tiny
        	\renewcommand\arraystretch{1.1}
        	\vskip -0.in
        	\caption{Comparison of different methods on classification accuracy (top 1). We adopt \textit{ResNet-18} as the backbones.}
        	\vskip -0.14in
        	\label{tab:resnet18}
        	\setlength{\tabcolsep}{4.3pt}
        	\begin{center}
        		\begin{small}
        			\begin{tabular}{llllll}
        				\toprule
        				\multicolumn{2}{c}{Model} & CIFAR-10 & CIFAR-100 & STL-10 & IN-200 \\ 
        				\hline
        				\hline
        				\multicolumn{2}{l}{Barlow Twins \cite{2021Barlow}} & 85.72 & 60.31 & 73.62 & 45.10 \\ \rowcolor{mygray}
        				\multicolumn{2}{c}{+ MetaMask$_{\textcolor{red}{+1.25}}$} & 87.53$_{\textcolor{red}{+1.81}}$ & 61.42$_{\textcolor{red}{+1.11}}$ & 73.97$_{\textcolor{red}{+0.35}}$ & 46.81$_{\textcolor{red}{+1.71}}$ \\ 
        				\hline
        				\hline
        				\text{SimCLR \cite{chen2020simple}} & & 81.73 & 58.27 & 70.09 & 44.46 \\ \rowcolor{mygray}
        				\multicolumn{2}{c}{+ MetaMask$_{\textcolor{red}{+3.32}}$} & 86.01$_{\textcolor{red}{+4.28}}$ & 61.03$_{\textcolor{red}{+2.76}}$ & 74.90$_{\textcolor{red}{+4.81}}$ & 45.87$_{\textcolor{red}{+1.41}}$ \\ 
        				\text{BYOL \cite{2020Bootstrap}} & & 88.05 & 60.94 & 72.04 & 46.72 \\ \rowcolor{mygray}
        				\multicolumn{2}{c}{+ MetaMask$_{\textcolor{red}{+0.28}}$} & 87.53$_{\textcolor{blue}{-0.52}}$ & 61.42$_{\textcolor{red}{+0.48}}$ & 73.97$_{\textcolor{red}{+1.93}}$ & 46.81$_{\textcolor{red}{+0.11}}$ \\ 
        				\text{DCL \cite{2020Debiased}} & & 85.61 & 59.29 & 71.18 & 45.77 \\ \rowcolor{mygray}
        				\multicolumn{2}{c}{+ MetaMask$_{\textcolor{red}{+1.65}}$} & 86.47$_{\textcolor{red}{+0.86}}$ & 60.82$_{\textcolor{red}{+1.53}}$ & 74.79$_{\textcolor{red}{+3.61}}$ & 46.35$_{\textcolor{red}{+0.58}}$ \\ 
        				\text{HCL \cite{2020Hard}} & & 85.27 & 61.21 & 71.92 & 46.90 \\ \rowcolor{mygray}
        				\multicolumn{2}{c}{+ MetaMask$_{\textcolor{red}{+0.95}}$} & 85.97$_{\textcolor{red}{+0.70}}$ & 61.63$_{\textcolor{red}{+0.42}}$ & 74.03$_{\textcolor{red}{+2.11}}$ & 46.06$_{\textcolor{red}{+0.55}}$ \\ 
        				\text{NNCLR \cite{2021Dwibedi}} & & 81.44 & 61.64 & 71.89 & 47.10 \\ \rowcolor{mygray}
        				\multicolumn{2}{c}{+ MetaMask$_{\textcolor{red}{+1.63}}$} & 85.78$_{\textcolor{red}{+4.34}}$ & 61.89$_{\textcolor{red}{+0.25}}$ & 74.02$_{\textcolor{red}{+2.13}}$ & 46.90$_{\textcolor{blue}{-0.20}}$ \\ 
        				\bottomrule
        			\end{tabular}
        		\end{small}
        	\end{center}
        	\vspace{-0.8cm}
		\end{minipage} \quad
		\begin{minipage}[t]{0.28\textwidth}
			\makeatletter\def\@captype{figure}
			\vskip 0.in
			\begin{center}
				\centerline{\includegraphics[width=1\columnwidth]{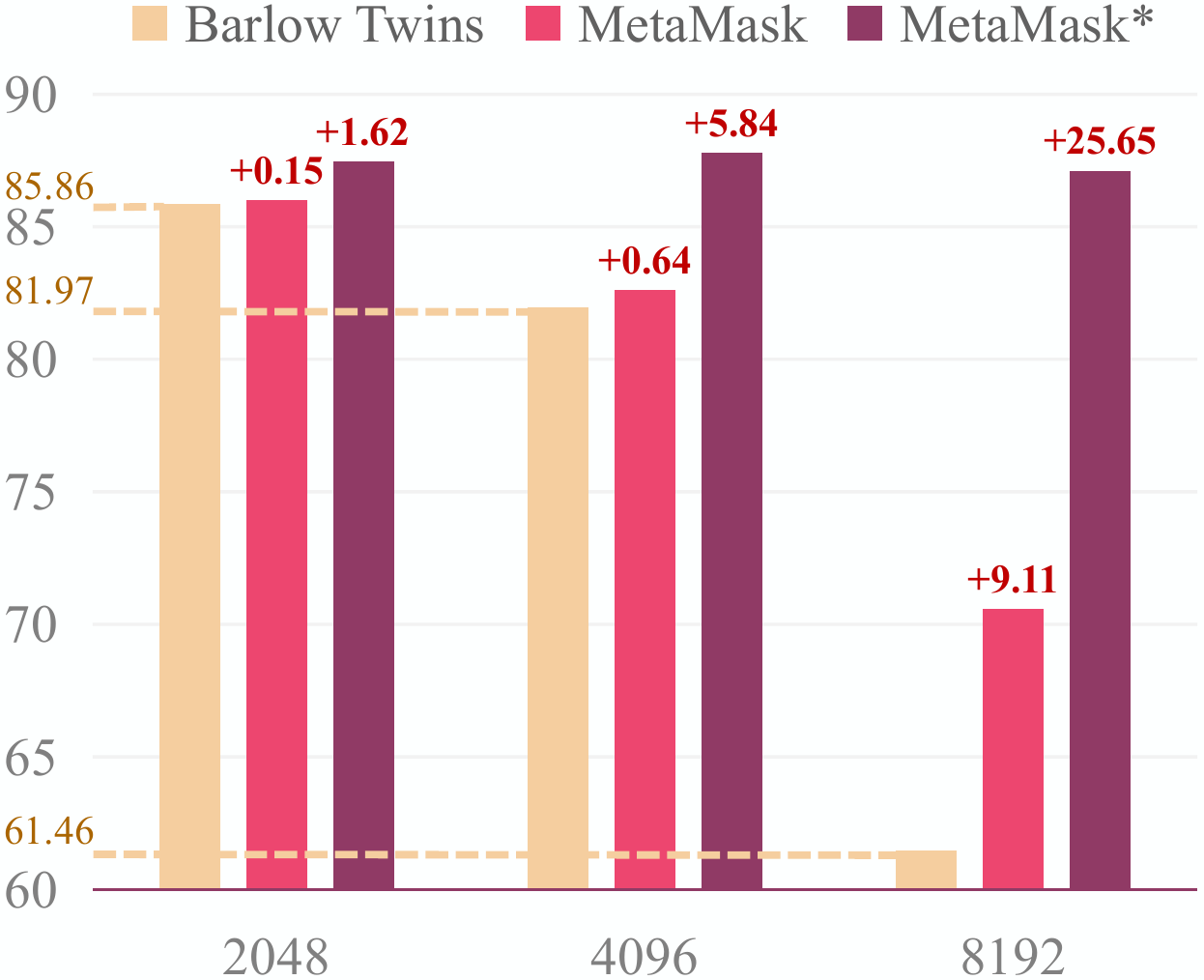}}
				\vskip -0.05in
				\caption{Comparisons of varying the dimensionality of the projection head on CIFAR-10 by using ResNet-18. $\ast$ denotes MetaMask using a trick of fixed learning rate instead of the cosine annealing strategy.}
				\label{fig:robust}
				\vskip -0.3in
			\end{center}
		\end{minipage}
	\end{minipage}
	\vspace{-0.15cm}
\end{table}

\section{Conclusion and Future Discussion} \label{sec:conclusion}
From the dimensional perspective, we first clarify that the dimensional redundancy and confounder issues are the essence of the current self-supervised learning paradigm's defects. Multiple motivating experiments are conducted to support our viewpoint, and the intriguing results are opposite to the traditional conclusions. To jointly tackle the dimensional redundancy and confounder issues, we propose a heuristic learning approach, called MetaMask, which leverages a second-derivative technique to learn a dimensional mask in order to adjust the gradient weights of different dimensions during training. The proposed theoretical analyses demonstrate that MetaMask obtains relatively tighter risk bounds for downstream classification compared to conventional contrastive methods. From the experiments, we observe a consistent improvement in the performance of our method over state-of-the-art methods on benchmark datasets.

One of our critical contributions is proposing a heuristic learning paradigm for self-supervised learning: regarding the dimensional redundancy reduction as a regularization term and further exploring to mask dimensions containing the confounder. In the current methodology, we train the learnable dimensional mask with the objective of improving the performance of masked representations on a specific \textit{self-supervised task}, i.e., contrastive learning, and provide the theoretical analysis to prove that such a learning approach can indeed prompt the model focus on acquiring discriminative information. However, contrastive learning is immutable for the \textit{self-supervised task} in MetaMask. It is commonly acknowledged that masked image modeling (MIM) methods \cite{bao2021beit, he2021masked, xie2021simmim} achieve impressive performance in self-supervised learning. Although the theoretical contribution bridging the gap between MIM loss and cross-entropy loss on downstream tasks has not yet emerged, we still deem combining MIM approaches with our proposed paradigm is attractive to the community.

\section*{Acknowledgements}
The authors would like to thank the anonymous reviewers for their valuable comments. This work is supported in part by National Natural Science Foundation of China No. 61976206 and No. 61832017, Key Special Project for Introduced Talents Team of Southern Marine Science and Engineering Guangdong Laboratory (Guangzhou) No. GML2019ZD0603, National Key Research and Development Program of China No. 2019YFB1405100, Foshan HKUST Projects (FSUST21-FYTRI01A, FSUST21-FYTRI02A), Beijing Outstanding Young Scientist Program NO. BJJWZYJH012019100020098, Beijing Academy of Artificial Intelligence (BAAI), the Fundamental Research Funds for the Central Universities, the Research Funds of Renmin University of China 21XNLG05, and Public Computing Cloud, Renmin University of China. This work is also supported in part by Intelligent Social Governance Platform, Major Innovation \& Planning Interdisciplinary Platform for the “Double-First Class” Initiative, Renmin University of China, and Public Policy and Decision-making Research Lab of Renmin University of China.

\bibliographystyle{unsrt}
\bibliography{reference}

%%%%%%%%%%%%%%%%%%%%%%%%%%%%%%%%%%%%%%%%%%%%%%%%%%%%%%%%%%%%
\section*{Checklist}

%%% BEGIN INSTRUCTIONS %%%
% The checklist follows the references. Please
% read the checklist guidelines carefully for information on how to answer these
% questions. For each question, change the default \answerTODO{} to \answerYes{},
% \answerNo{}, or \answerNA{}. You are strongly encouraged to include a {\bf
% justification to your answer}, either by referencing the appropriate section of
% your paper or providing a brief inline description. For example:
% \begin{itemize}
%   \item Did you include the license to the code and datasets? \answerYes{See Section~\ref{gen_inst}.}
%   \item Did you include the license to the code and datasets? \answerNo{The code and the data are proprietary.}
%   \item Did you include the license to the code and datasets? \answerNA{}
% \end{itemize}
% Please do not modify the questions and only use the provided macros for your
% answers.  Note that the Checklist section does not count towards the page
% limit.  In your paper, please delete this instructions block and only keep the
% Checklist section heading above along with the questions/answers below.
%%% END INSTRUCTIONS %%%

\begin{enumerate}

\item For all authors...
\begin{enumerate}
  \item Do the main claims made in the abstract and introduction accurately reflect the paper's contributions and scope?
    \answerYes{}
  \item Did you describe the limitations of your work?
    \answerYes{See Section \ref{sec:conclusion}.}
  \item Did you discuss any potential negative societal impacts of your work?
    \answerNo{Since this work presents a general technique to tackle the dimensional confounder for self-supervised learning, we did not see any particular foreseeable ethical consequence.}
  \item Have you read the ethics review guidelines and ensured that your paper conforms to them?
    \answerYes{}
\end{enumerate}

\item If you are including theoretical results...
\begin{enumerate}
  \item Did you state the full set of assumptions of all theoretical results?
    \answerYes{See Section \ref{sec:theory}.}
        \item Did you include complete proofs of all theoretical results?
    \answerYes{See Appendix \ref{app:discusion} and Appendix \ref{app:proof}.}
\end{enumerate}

\item If you ran experiments...
\begin{enumerate}
  \item Did you include the code, data, and instructions needed to reproduce the main experimental results (either in the supplemental material or as a URL)?
    \answerYes{See the supplementary files.}
  \item Did you specify all the training details (e.g., data splits, hyper-parameters, how they were chosen)?
    \answerYes{See Section \ref{sec:experiments}, Appendix \ref{app:extendedexp}, and the supplementary files.}
        \item Did you report error bars (e.g., with respect to the random seed after running experiments multiple times)?
    \answerNo{The experimental results provided by benchmark methods do not contain such error bars. We also follow this principle, and we reckon the reason behind such a behavior is that the comparisons are conducted in balanced datasets, so the difference between results is not excessively large.}
        \item Did you include the total amount of compute and the type of resources used (e.g., type of GPUs, internal cluster, or cloud provider)?
    \answerYes{See Section \ref{sec:intro} and Appendix \ref{app:extendedexp}.}
\end{enumerate}

\item If you are using existing assets (e.g., code, data, models) or curating/releasing new assets...
\begin{enumerate}
  \item If your work uses existing assets, did you cite the creators?
    \answerYes{See Section \ref{sec:intro}, Section \ref{sec:relatedwork}, Section \ref{sec:theory}, and Section \ref{sec:experiments}.}
  \item Did you mention the license of the assets?
    \answerNo{The assets we adopted are totally established and public to the community.}
  \item Did you include any new assets either in the supplemental material or as a URL?
    \answerYes{See the supplementary files.}
  \item Did you discuss whether and how consent was obtained from people whose data you're using/curating?
    \answerNo{The datasets are all public to the community.}
  \item Did you discuss whether the data you are using/curating contains personally identifiable information or offensive content?
    \answerNo{The datasets have been discussed, validated, and used by many researchers.}
\end{enumerate}

\item If you used crowdsourcing or conducted research with human subjects...
\begin{enumerate}
  \item Did you include the full text of instructions given to participants and screenshots, if applicable?
    \answerNA{}
  \item Did you describe any potential participant risks, with links to Institutional Review Board (IRB) approvals, if applicable?
    \answerNA{}
  \item Did you include the estimated hourly wage paid to participants and the total amount spent on participant compensation?
    \answerNA{}
\end{enumerate}

\end{enumerate}

%%%%%%%%%%%%%%%%%%%%%%%%%%%%%%%%%%%%%%%%%%%%%%%%%%%%%%%%%%%%

\clearpage
\appendix

\section{Appendix}

\subsection{Elaborating MetaMask's Behavior of Masking Gradient Contribution} \label{app:discusion}
For better understanding the intuition behind the behavior of the proposed MetaMask, we elaborate the procedure of our method masking gradient contributions of dimensions. We restate Equation \ref{eq:cl} of contrastive loss by converting the original projected features $\boldsymbol{z}$ into the masked feature $\tilde{\boldsymbol{z}}$:
\begin{equation} \label{eq:maskedcl}
    \begin{aligned}
        \mathcal{L}_{maskedcl} &= - \sum \sum \log \frac{\exp\left[d\left(\tilde{\boldsymbol{z}}, \tilde{\boldsymbol{z}}^+ \right) / \tau\right]}{\sum \sum \mathbb{1} \cdot \exp\left[d\left(\tilde{\boldsymbol{z}}, \tilde{\boldsymbol{z}}^\prime \right) / \tau\right]} \\
        &= - \sum \sum \log \frac{\exp\left[d\left(g^{cl}_{\vartheta}\left(\tilde{\boldsymbol{h}}\right), g^{cl}_{\vartheta}\left(\tilde{\boldsymbol{h}}^+ \right) \right) / \tau\right]}{\sum \sum \mathbb{1} \cdot \exp\left[d\left(g^{cl}_{\vartheta}\left(\tilde{\boldsymbol{h}}\right), g^{cl}_{\vartheta}\left(\tilde{\boldsymbol{h}}^\prime\right) \right) / \tau\right]} \\
        &= - \sum \sum \log \frac{\exp\left[d\left(g^{cl}_{\vartheta}\left({\boldsymbol{h}} \otimes \mathcal{M}\right), g^{cl}_{\vartheta}\left({\boldsymbol{h}}^+ \otimes \mathcal{M}\right) \right) / \tau\right]}{\sum \sum \mathbb{1} \cdot \exp\left[d\left(g^{cl}_{\vartheta}\left({\boldsymbol{h}}\otimes \mathcal{M}\right), g^{cl}_{\vartheta}\left({\boldsymbol{h}}^\prime\otimes \mathcal{M}\right) \right) / \tau\right]}, 
    \end{aligned}
\end{equation}
which is derived by considering Equation \ref{eq:maskh}. To simplify the equation, we hold
\begin{equation}
	{\mathcal{L}_{maskedcl}} = \sum_{i \in \llbracket 1, N\rrbracket} \mathop{\sum_{j \in \llbracket 1, M - 1\rrbracket}}\limits_{j^+ \in \llbracket j + 1, M\rrbracket} \mathcal{L}_{i, j, j^+}.
	\label{eq:simmaskedcl}
\end{equation}
Therefore, the indicator function $\mathbb{1}_{\left[ i \neq i^\prime \lor j \neq j^\prime \right]}$ can be substituted as follows:
\begin{equation} \label{eq:simmaskedcl1}
    \mathcal{L}_{i, j, j^+} = -\log \frac{\exp\left[d\left(g\left({\boldsymbol{h}} \otimes \mathcal{M}\right), g\left({\boldsymbol{h}}^+ \otimes \mathcal{M}\right) \right) / \tau\right]}{\exp\left[d\left(g\left({\boldsymbol{h}} \otimes \mathcal{M}\right), g\left({\boldsymbol{h}}^+ \otimes \mathcal{M}\right) \right) / \tau\right] + \mathop{\sum}\limits_{\boldsymbol{h}^-} \exp\left[d\left(g\left({\boldsymbol{h}}\otimes \mathcal{M}\right), g\left({\boldsymbol{h}}^-\otimes \mathcal{M}\right) \right) / \tau\right]}, 
\end{equation}
where $g\left(\cdot\right)$ denotes $g^{cl}_{\vartheta}\left(\cdot\right)$, and $\boldsymbol{h}^-$ is neither equal to $\boldsymbol{h}$ nor equal to $\boldsymbol{h}^+$.
\begin{proposition}
	The mask $\mathcal{M}$ exists on each partial differential equation of the gradient of $\mathcal{L}_{i, j, j^+}$:
	\label{pro:mexist}
	\begin{equation} \label{eq:mexisth}
	    \begin{aligned}
	        \nabla_{\boldsymbol{h}}\mathcal{L}_{i, j, j^+} = &\frac{\mathop{\sum}\limits_{\boldsymbol{h}^-} \exp\left[d\left(g\left({\boldsymbol{h}}\otimes \mathcal{M}\right), g\left({\boldsymbol{h}}^-\otimes \mathcal{M}\right) \right) / \tau\right] - \exp\left[d\left(g\left({\boldsymbol{h}} \otimes \mathcal{M}\right), g\left({\boldsymbol{h}}^+ \otimes \mathcal{M}\right) \right) / \tau\right]}{\mathop{\sum}\limits_{\boldsymbol{h}^-} \exp\left[d\left(g\left({\boldsymbol{h}}\otimes \mathcal{M}\right), g\left({\boldsymbol{h}}^-\otimes \mathcal{M}\right) \right) / \tau\right]} \\
	        & \times \left[ \mathop{\sum}\limits_{{\boldsymbol{h}^-}^\prime} \frac{\exp\left[d\left(g\left({\boldsymbol{h}}\otimes \mathcal{M}\right), g\left({{\boldsymbol{h}}^-}^\prime\otimes \mathcal{M}\right) \right) / \tau\right]}{\mathop{\sum}\limits_{\boldsymbol{h}^-} \exp\left[d\left(g\left({\boldsymbol{h}}\otimes \mathcal{M}\right), g\left({\boldsymbol{h}}^-\otimes \mathcal{M}\right) \right) / \tau\right]} \times g\left({{\boldsymbol{h}}^-}^\prime\otimes \mathcal{M}\right) - g\left({\boldsymbol{h}}^+ \otimes \mathcal{M}\right)\right] \\
	        & \times \mathop{\prod}\limits_{k \in \llbracket 1, N^g \rrbracket} \boldsymbol{w}_k \times \nabla_{\boldsymbol{h}}\boldsymbol{h} \otimes \mathcal{M}
	   \end{aligned}
	\end{equation}
	\begin{equation} \label{eq:mexisthp}
	    \begin{aligned}
	        \nabla_{\boldsymbol{h}^+}\mathcal{L}_{i, j, j^+} = &\frac{\exp\left[d\left(g\left({\boldsymbol{h}} \otimes \mathcal{M}\right), g\left({\boldsymbol{h}}^+ \otimes \mathcal{M}\right) \right) / \tau\right] - \mathop{\sum}\limits_{\boldsymbol{h}^-} \exp\left[d\left(g\left({\boldsymbol{h}}\otimes \mathcal{M}\right), g\left({\boldsymbol{h}}^-\otimes \mathcal{M}\right) \right) / \tau\right]}{\mathop{\sum}\limits_{\boldsymbol{h}^-} \exp\left[d\left(g\left({\boldsymbol{h}}\otimes \mathcal{M}\right), g\left({\boldsymbol{h}}^-\otimes \mathcal{M}\right) \right) / \tau\right]} \\
	        &\times g\left({\boldsymbol{h}} \otimes \mathcal{M}\right) \times \mathop{\prod}\limits_{k \in \llbracket 1, N^g \rrbracket} \boldsymbol{w}_k \times \nabla_{\boldsymbol{h}^+}\boldsymbol{h}^+ \otimes \mathcal{M}
        \end{aligned}
	\end{equation}
	\begin{equation} \label{eq:mexisthn}
	    \begin{aligned}
	        \nabla_{\boldsymbol{h}^-}\mathcal{L}_{i, j, j^+} = &\frac{\mathop{\sum}\limits_{\boldsymbol{h}^-} \exp\left[d\left(g\left({\boldsymbol{h}}\otimes \mathcal{M}\right), g\left({\boldsymbol{h}}^-\otimes \mathcal{M}\right) \right) / \tau\right] - \exp\left[d\left(g\left({\boldsymbol{h}} \otimes \mathcal{M}\right), g\left({\boldsymbol{h}}^+ \otimes \mathcal{M}\right) \right) / \tau\right]}{\mathop{\sum}\limits_{\boldsymbol{h}^-} \exp\left[d\left(g\left({\boldsymbol{h}}\otimes \mathcal{M}\right), g\left({\boldsymbol{h}}^-\otimes \mathcal{M}\right) \right) / \tau\right]} \\
	        &\times \frac{\exp\left[d\left(g\left({\boldsymbol{h}}\otimes \mathcal{M}\right), g\left({\boldsymbol{h}}^-\otimes \mathcal{M}\right) \right) / \tau\right]}{\mathop{\sum}\limits_{{\boldsymbol{h}^-}^\prime} \exp\left[d\left(g\left({\boldsymbol{h}}\otimes \mathcal{M}\right), g\left({{\boldsymbol{h}}^-}^\prime \otimes \mathcal{M}\right) \right) / \tau\right]} \times g\left( \boldsymbol{h} \otimes \mathcal{M} \right) \times \mathop{\prod}\limits_{k \in \llbracket 1, N^g \rrbracket} \boldsymbol{w}_k \\
	        &\times \nabla_{\boldsymbol{h}^-}\boldsymbol{h}^- \otimes \mathcal{M}
	    \end{aligned}
	\end{equation}
	where $\left\{\boldsymbol{w}_k \big| k \in \llbracket 1, N^g \rrbracket \right\}$ denotes the weights of the projection head $g\left( \cdot \right)$, $N^g$ denotes the number of corresponding projection layers, and ${\boldsymbol{h}^-}^\prime$ is neither equal to $\boldsymbol{h}$ nor equal to $\boldsymbol{h}^+$ (note that $\boldsymbol{h}^-$ and ${\boldsymbol{h}^-}^\prime$ are two independent variables). $\mathcal{M}$ is the dimensional mask.
\end{proposition}
Proposition \ref{pro:mexist} states that the dimensional mask $\mathcal{M}$ consistently exists on each partial differential equation of the gradient of the contrastive loss $\mathcal{L}_{i, j, j^+}$. We further observe that there exists 7 types of irreducible terms containing $\mathcal{M}$: $g\left({\boldsymbol{h}}\otimes \mathcal{M}\right)$, $g\left({\boldsymbol{h}^+}\otimes \mathcal{M}\right)$, $g\left({\boldsymbol{h}^-}\otimes \mathcal{M}\right)$, $g\left({\boldsymbol{h}^-}^\prime\otimes \mathcal{M}\right)$, $\nabla_{\boldsymbol{h}}\boldsymbol{h} \otimes \mathcal{M}$, $\nabla_{\boldsymbol{h}^+}\boldsymbol{h} \otimes \mathcal{M}$, $\nabla_{\boldsymbol{h}^-}\boldsymbol{h} \otimes \mathcal{M}$, where the first 4 types are in the regular computing manner so that $\mathcal{M}$ can directly change the values of different dimensions by assigning dimension-specific weights, and the last 3 types are the first-order derivatives. Note that the scalar derivation, e.g., the loss, of a matrix can be thought of as derivation of each element of the matrix, and such a derivation is derived and then putting the element-wise results in the order of the matrix to get a gradient matrix of the same dimension.

Following such a law, we utilize Chain Rules to derive the derivative of the loss $\mathcal{L}_{i, j, j^+}$ with respect to each element in the batch matrix containing $\boldsymbol{h}$, $\boldsymbol{h}^+$, $\boldsymbol{h}^-$ and further obtain the derivative matrix:
\begin{equation}
    \begin{aligned}
        &\frac{\partial \mathcal{L}_{i, j, j^+}}{\partial \mathcal{X}^\prime_{p, q}} = \frac{\partial \mathcal{L}_{i, j, j^+}}{\partial \mathcal{B}_{p, q}}\frac{\partial \mathcal{B}_{p, q}}{\partial \mathcal{X}^\prime_{p, q}} = \frac{\partial \mathcal{L}_{i, j, j^+}}{\partial\left[\left(\mathcal{B} \otimes \hat{\mathcal{M}}\right)_{p, q}\right]} \frac{\partial\left[\left(\mathcal{B} \otimes \hat{\mathcal{M}}\right)_{p, q}\right]}{\partial \mathcal{B}_{p, q}}\frac{\partial \mathcal{B}_{p, q}}{\partial \mathcal{X}^\prime_{p, q}} \\
        &= \left[ {\begin{array}{*{20}{c}}
	        \boldsymbol{\omega_1}\frac{\partial\lfloor f_{\theta}\left(\boldsymbol{x}\right) \rceil^1}{\partial\boldsymbol{x}}&\boldsymbol{\omega_2}\frac{\partial\lfloor f_{\theta}\left(\boldsymbol{x}\right) \rceil^2}{\partial\boldsymbol{x}}&\cdots&\boldsymbol{\omega_{D-1}}\frac{\partial\lfloor f_{\theta}\left(\boldsymbol{x}\right) \rceil^{D-1}}{\partial\boldsymbol{x}}&\boldsymbol{\omega_D}\frac{\partial\lfloor f_{\theta}\left(\boldsymbol{x}\right) \rceil^D}{\partial\boldsymbol{x}}\\
	        \boldsymbol{\omega_1}\frac{\partial\lfloor f_{\theta}\left(\boldsymbol{x}^+\right) \rceil^1}{\partial\boldsymbol{x}^+}&\boldsymbol{\omega_2}\frac{\partial\lfloor f_{\theta}\left(\boldsymbol{x}^+\right) \rceil^2}{\partial\boldsymbol{x}^+}&\cdots&\boldsymbol{\omega_{D-1}}\frac{\partial\lfloor f_{\theta}\left(\boldsymbol{x}^+\right) \rceil^{D-1}}{\partial\boldsymbol{x}^+}&\boldsymbol{\omega_D}\frac{\partial\lfloor f_{\theta}\left(\boldsymbol{x}^+\right) \rceil^D}{\partial\boldsymbol{x}^+}\\
	        \boldsymbol{\omega_1}\frac{\partial\lfloor f_{\theta}\left(\boldsymbol{x}^-_1\right) \rceil^1}{\partial\boldsymbol{x}^-_1}&\boldsymbol{\omega_2}\frac{\partial\lfloor f_{\theta}\left(\boldsymbol{x}^-_1\right) \rceil^2}{\partial\boldsymbol{x}^-_1}&\cdots&\boldsymbol{\omega_{D-1}}\frac{\partial\lfloor f_{\theta}\left(\boldsymbol{x}^-_1\right) \rceil^{D-1}}{\partial\boldsymbol{x}^-_1}&\boldsymbol{\omega_D}\frac{\partial\lfloor f_{\theta}\left(\boldsymbol{x}^-_1\right) \rceil^D}{\partial\boldsymbol{x}^-_1}\\
	        \boldsymbol{\omega_1}\frac{\partial\lfloor f_{\theta}\left(\boldsymbol{x}^-_2\right) \rceil^1}{\partial\boldsymbol{x}^-_2}&\boldsymbol{\omega_2}\frac{\partial\lfloor f_{\theta}\left(\boldsymbol{x}^-_2\right) \rceil^2}{\partial\boldsymbol{x}^-_2}&\cdots&\boldsymbol{\omega_{D-1}}\frac{\partial\lfloor f_{\theta}\left(\boldsymbol{x}^-_2\right) \rceil^{D-1}}{\partial\boldsymbol{x}^-_2}&\boldsymbol{\omega_D}\frac{\partial\lfloor f_{\theta}\left(\boldsymbol{x}^-_2\right) \rceil^D}{\partial\boldsymbol{x}^-_2}\\
	        \cdots&\cdots&\cdots&\cdots&\cdots\\
	        \boldsymbol{\omega_1}\frac{\partial\lfloor f_{\theta}\left(\boldsymbol{x}^-_{N-3}\right) \rceil^1}{\partial\boldsymbol{x}^-_{N-3}}&\boldsymbol{\omega_2}\frac{\partial\lfloor f_{\theta}\left(\boldsymbol{x}^-_{N-3}\right) \rceil^2}{\partial\boldsymbol{x}^-_{N-3}}&\cdots&\boldsymbol{\omega_{D-1}}\frac{\partial\lfloor f_{\theta}\left(\boldsymbol{x}^-_{N-3}\right) \rceil^{D-1}}{\partial\boldsymbol{x}^-_{N-3}}&\boldsymbol{\omega_D}\frac{\partial\lfloor f_{\theta}\left(\boldsymbol{x}^-_{N-3}\right) \rceil^D}{\partial\boldsymbol{x}^-_{N-3}}\\
	        \boldsymbol{\omega_1}\frac{\partial\lfloor f_{\theta}\left(\boldsymbol{x}^-_{N-2}\right) \rceil^1}{\partial\boldsymbol{x}^-_{N-2}}&\boldsymbol{\omega_2}\frac{\partial\lfloor f_{\theta}\left(\boldsymbol{x}^-_{N-2}\right) \rceil^2}{\partial\boldsymbol{x}^-_{N-2}}&\cdots&\boldsymbol{\omega_{D-1}}\frac{\partial\lfloor f_{\theta}\left(\boldsymbol{x}^-_{N-2}\right) \rceil^{D-1}}{\partial\boldsymbol{x}^-_{N-2}}&\boldsymbol{\omega_D}\frac{\partial\lfloor f_{\theta}\left(\boldsymbol{x}^-_{N-2}\right) \rceil^D}{\partial\boldsymbol{x}^-_{N-2}}\\
	    \end{array}} \right]
    \end{aligned}
\end{equation}
where $\mathcal{B} = \left[\boldsymbol{h}, \boldsymbol{h}^+, \boldsymbol{h}^-_1, \boldsymbol{h}^-_2, \ldots, \boldsymbol{h}^-_{N - 3}, \boldsymbol{h}^-_{N - 2}\right]$ is a batch of features with $N$ examples, and $\mathcal{X}^\prime$ is the corresponding input data matrix. $\boldsymbol{h}$, $\boldsymbol{h}^+$, and $\boldsymbol{h}^-$ are feature vectors containing $D$ dimensions. $\hat{\mathcal{M}} = \left[\mathcal{M}, \mathcal{M}, \ldots, \mathcal{M}, \mathcal{M}\right]$ is an extended matrix containing $N$ copies of $\mathcal{M}$. Intriguingly, we observe that $\mathcal{M}$ generally assign the dimension-specific weights for each samples, which empowers MetaMask to naturally reduce the gradient effect of specific dimensions containing the confounder by employing the meta-learning paradigm with the objective of improving the performance of masked representations on a typical self-supervised task. After clarifying the dimension-specific weighting mechanism from the perspective of gradient analysis, we further demonstrate the effectiveness of MetaMask's training paradigm as follows.

\subsection{Proofs} \label{app:proof}
In order to prove Theorem \ref{the:tighter} and the conclusion that the bounds of supervised cross-entropy loss obtained by MetaMask are tighter than typical contrastive learning methods, we provide separated proofs for the corresponding two parts of Theorem \ref{the:tighter} and the tighter bounds on downstream cross-entropy loss.

\subsubsection{Proof for the Equality Part} \label{prf:equal}
\begin{figure}
	\centering
	\includegraphics[width=0.9\textwidth]{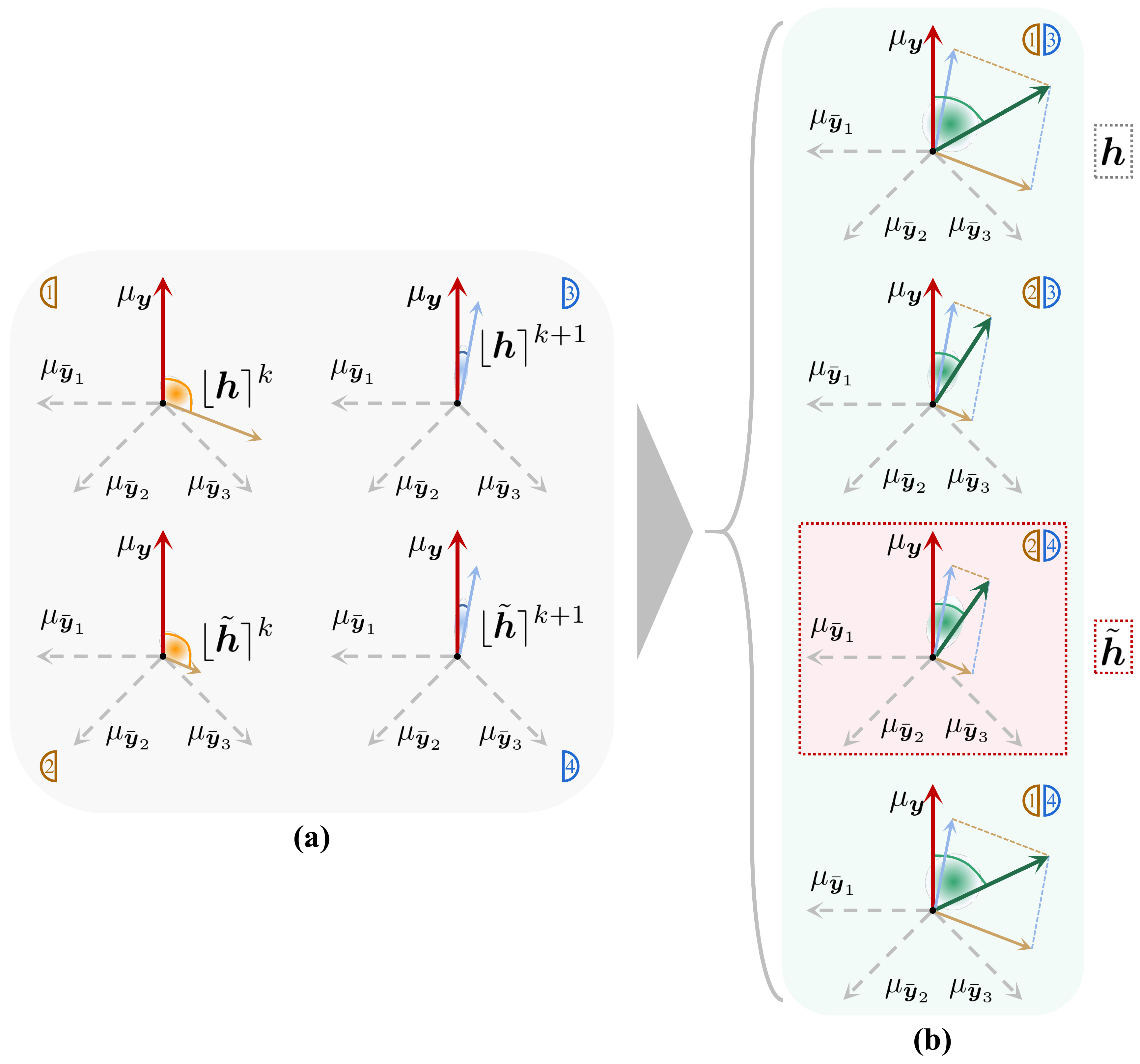}
	\vskip -0.1in
	\caption{Evidence examples of the proofs for Theorem \ref{the:tighter}. (a) the visual examples of the $k$-th dimension of the unmasked representation $\boldsymbol{h}$ or the masked representation $\tilde{\boldsymbol{h}}$. (b) the real and hypothetical examples of $\boldsymbol{h}$ or $\tilde{\boldsymbol{h}}$. $\mu_{\boldsymbol{y}}$ denotes the target category's high-dimensional feature, and $\left\{\mu_{\bar{\boldsymbol{y}_1}}, \mu_{\bar{\boldsymbol{y}_2}}, \mu_{\bar{\boldsymbol{y}_3}}\right\}$ denote the features of other categories.}
	\label{fig:proofexample}
	\vspace{-0.5cm}
\end{figure}

To prove
\begin{equation} \label{eq:equterm}
	\Phi\left(\lfloor g^{cl}_{\vartheta}\left(\tilde{\boldsymbol{h}}\right)\rceil^k \Big| \boldsymbol{y}\right) \cong \Phi\left(\lfloor g^{cl}_{\vartheta}\left({\boldsymbol{h}}\right)\rceil^k \Big| \boldsymbol{y}\right),
\end{equation}
we observe that the only variable that differs on both sides of the Equation \ref{eq:equterm} is the representation, which is masked on the left hand side of the equation while unmasked on the right hand side of the equation. Then, we substitute the definition of $\Phi\left( \cdot \right)$ into Equation \ref{eq:equterm} and derive
\begin{equation} \label{eq:extendequterm}
    \begin{aligned}
        \mathbb{E}_{\mathcal{P}\left(y\right)} \Bigg[ \mathbb{E}_{\mathcal{P}\left(\lfloor g^{cl}_{\vartheta}\left(\tilde{\boldsymbol{h}}\right)\rceil^k | y\right)}&\delta\left( \lfloor g^{cl}_{\vartheta}\left(\tilde{\boldsymbol{h}}\right)\rceil^k, \mathbb{E}_{\mathcal{P}\left(\lfloor g^{cl}_{\vartheta}\left(\tilde{\boldsymbol{h}}\right)\rceil^k | y\right)} \right) \Bigg] \\ \cong \mathbb{E}_{\mathcal{P}\left(y\right)} \Bigg[ \mathbb{E}_{\mathcal{P}\left(\lfloor g^{cl}_{\vartheta}\left({\boldsymbol{h}}\right)\rceil^k | y\right)}&\delta\left( \lfloor g^{cl}_{\vartheta}\left({\boldsymbol{h}}\right)\rceil^k, \mathbb{E}_{\mathcal{P}\left(\lfloor g^{cl}_{\vartheta}\left({\boldsymbol{h}}\right)\rceil^k | y\right)} \right) \Bigg].
    \end{aligned}
\end{equation}
In our approach, the learned representation generally exists in high-dimensional embedding space, e.g., 1024, 2048, 8192, etc. Therefore, in high-dimensional embedding space, the left hand side of the equation can be extended to
\begin{equation} \label{eq:extendequterm2}
    \begin{aligned}
        &\mathbb{E}_{\mathcal{P}\left(y\right)}\left[\mathbb{E}_{\mathcal{P}\left(\lfloor g^{cl}_{\vartheta}\left(\tilde{\boldsymbol{h}}\right)\rceil^k | y\right)}\delta\left( \lfloor g^{cl}_{\vartheta}\left(\tilde{\boldsymbol{h}}\right)\rceil^k, \mathbb{E}_{\mathcal{P}\left(\lfloor g^{cl}_{\vartheta}\left(\tilde{\boldsymbol{h}}\right)\rceil^k | y\right)} \right)\right] \\
        = \ & \mathbb{E}_{\mathcal{P}\left(y\right)}\left[\mathbb{E}_{\mathcal{P}\left(\lfloor g^{cl}_{\vartheta}\left(\tilde{\boldsymbol{h}}\right)\rceil^k | y\right)} - \log\left(\cos\left( \lfloor g^{cl}_{\vartheta}\left(\tilde{\boldsymbol{h}}\right)\rceil^k, \mathbb{E}_{\mathcal{P}\left(\lfloor g^{cl}_{\vartheta}\left(\tilde{\boldsymbol{h}}\right)\rceil^k | y\right)} \right)\right)\right] \\
        = \ & \mathbb{E}_{\mathcal{P}\left(y\right)}\left[\mathbb{E}_{\mathcal{P}\left(\lfloor g^{cl}_{\vartheta}\left(\tilde{\boldsymbol{h}}\right)\rceil^k | y\right)} - \log\frac{\lfloor g^{cl}_{\vartheta}\left(\tilde{\boldsymbol{h}}\right)\rceil^k \times \mathbb{E}_{\mathcal{P}\left(\lfloor g^{cl}_{\vartheta}\left(\tilde{\boldsymbol{h}}\right)\rceil^k | y\right)}}{||\lfloor g^{cl}_{\vartheta}\left(\tilde{\boldsymbol{h}}\right)\rceil^k|| \times ||\mathbb{E}_{\mathcal{P}\left(\lfloor g^{cl}_{\vartheta}\left(\tilde{\boldsymbol{h}}\right)\rceil^k | y\right)}||}\right] \\
        = \ & \mathbb{E}_{\mathcal{P}\left(\lfloor g^{cl}_{\vartheta}\left(\tilde{\boldsymbol{h}}\right)\rceil^k, y\right)} - \log\frac{\lfloor g^{cl}_{\vartheta}\left(\tilde{\boldsymbol{h}}\right)\rceil^k \times \mu_{\boldsymbol{y}}}{||\lfloor g^{cl}_{\vartheta}\left(\tilde{\boldsymbol{h}}\right)\rceil^k|| \times ||\mu_{\boldsymbol{y}}||} \\
        \approx \ & \mathbb{E}_{\mathcal{P}\left(\lfloor \tilde{\boldsymbol{h}} \rceil^k, y\right)} - \log\frac{\lfloor \tilde{\boldsymbol{h}} \rceil^k \times \mu_{\boldsymbol{y}}}{||\lfloor \tilde{\boldsymbol{h}} \rceil^k|| \times ||\mu_{\boldsymbol{y}}||} \\
        = \ & - \mathbb{E}_{\mathcal{P}\left(\lfloor \tilde{\boldsymbol{h}} \rceil^k, y\right)}\log\frac{\lfloor \tilde{\boldsymbol{h}} \rceil^k \times \mu_{\boldsymbol{y}}}{||\lfloor \tilde{\boldsymbol{h}} \rceil^k|| \times ||\mu_{\boldsymbol{y}}||},
    \end{aligned}
\end{equation}
which is derived by following the Bayes Rule, the definition of $\delta\left(\cdot, \cdot\right)$ in Theorem \ref{the:connect}, and the fact that $g^{cl}_{\vartheta}\left(\cdot\right)$ is an MLP without amounts of deep layers so that the embedding space shift is limited. Note that $\mu_{\boldsymbol{y}}$ denotes the center of categories' feature or the corresponding classification weight of the classifier for a specific category, and $\boldsymbol{y} = \left\{ \boldsymbol{y}_l \big| l \in \llbracket {1, N^L} \rrbracket \right\}$, where $N^L$ is the number of categories. Then, we bring Equation \ref{eq:extendequterm2} into Equation \ref{eq:extendequterm} and get
\begin{equation} \label{eq:extendequterm3}
    \begin{aligned}
        - \mathbb{E}_{\mathcal{P}\left(\lfloor \tilde{\boldsymbol{h}} \rceil^k, y\right)}\log\frac{\lfloor \tilde{\boldsymbol{h}} \rceil^k \times \mu_{\boldsymbol{y}}}{||\lfloor \tilde{\boldsymbol{h}} \rceil^k|| \times ||\mu_{\boldsymbol{y}}||} &\cong - \mathbb{E}_{\mathcal{P}\left(\lfloor {\boldsymbol{h}} \rceil^k, y\right)}\log\frac{\lfloor {\boldsymbol{h}} \rceil^k \times \mu_{\boldsymbol{y}}}{||\lfloor {\boldsymbol{h}} \rceil^k|| \times ||\mu_{\boldsymbol{y}}||} \\
        \mathbb{E}_{\mathcal{P}\left(\lfloor \tilde{\boldsymbol{h}} \rceil^k, y\right)}\log\frac{\lfloor \tilde{\boldsymbol{h}} \rceil^k \times \mu_{\boldsymbol{y}}}{||\lfloor \tilde{\boldsymbol{h}} \rceil^k|| \times ||\mu_{\boldsymbol{y}}||} &\cong \mathbb{E}_{\mathcal{P}\left(\lfloor {\boldsymbol{h}} \rceil^k, y\right)}\log\frac{\lfloor {\boldsymbol{h}} \rceil^k \times \mu_{\boldsymbol{y}}}{||\lfloor {\boldsymbol{h}} \rceil^k|| \times ||\mu_{\boldsymbol{y}}||}.
    \end{aligned}
\end{equation}
To prove Equation \ref{eq:extendequterm3}, we demonstrate an evidence example in Figure \ref{fig:proofexample}. As shown in the subfigure (a), the cosine error risk of the target vector $\mu_{\boldsymbol{y}}$ and the candidate vector, e.g., $\lfloor \boldsymbol{h} \rceil^k$, $\lfloor \tilde{\boldsymbol{h}} \rceil^k$, etc, is insensitive to the specific value of vectors, i.e., the cosine error risk keeps generally consistent as the values of vectors fall or vice versa. Such an assumption is supported by comparing plot (1 with plot (2 or comparing plot 3) with plot 4). Therefore, we derive
\begin{equation} \label{eq:extendequterm4}
    \log\frac{\lfloor \tilde{\boldsymbol{h}} \rceil^k \times \mu_{\boldsymbol{y}}}{||\lfloor \tilde{\boldsymbol{h}} \rceil^k|| \times ||\mu_{\boldsymbol{y}}||} = \log\frac{\lfloor {\boldsymbol{h}} \rceil^k \times \mu_{\boldsymbol{y}}}{||\lfloor {\boldsymbol{h}} \rceil^k|| \times ||\mu_{\boldsymbol{y}}||}
\end{equation}
so that
\begin{equation} \label{eq:extendequterm5}
    \mathbb{E}_{\mathcal{P}\left(\lfloor \tilde{\boldsymbol{h}} \rceil^k, y\right)}\log\frac{\lfloor \tilde{\boldsymbol{h}} \rceil^k \times \mu_{\boldsymbol{y}}}{||\lfloor \tilde{\boldsymbol{h}} \rceil^k|| \times ||\mu_{\boldsymbol{y}}||} \cong \mathbb{E}_{\mathcal{P}\left(\lfloor {\boldsymbol{h}} \rceil^k, y\right)}\log\frac{\lfloor {\boldsymbol{h}} \rceil^k \times \mu_{\boldsymbol{y}}}{||\lfloor {\boldsymbol{h}} \rceil^k|| \times ||\mu_{\boldsymbol{y}}||}.
\end{equation}

\subsubsection{Proof for the Inequality Part} \label{prf:inequal}
To prove
\begin{equation} \label{eq:inequterm}
	\Phi\left(g^{cl}_{\vartheta}\left(\tilde{\boldsymbol{h}}\right) \Big| \boldsymbol{y}\right) \leq \Phi\left(g^{cl}_{\vartheta}\left({\boldsymbol{h}}\right) \Big| \boldsymbol{y}\right)
\end{equation}
we follow the formula transformation of Equation \ref{eq:extendequterm2} and Equation \ref{eq:extendequterm3} to derive
\begin{equation} \label{eq:extendinequterm}
    \begin{aligned}
        \mathbb{E}_{\mathcal{P}\left(y\right)}\left[\mathbb{E}_{\mathcal{P}\left(g^{cl}_{\vartheta}\left(\tilde{\boldsymbol{h}}\right) | y\right)}\delta\left( g^{cl}_{\vartheta}\left(\tilde{\boldsymbol{h}}\right), \mathbb{E}_{\mathcal{P}\left(g^{cl}_{\vartheta}\left(\tilde{\boldsymbol{h}}\right) | y\right)} \right)\right] &\leq \mathbb{E}_{\mathcal{P}\left(y\right)}\left[\mathbb{E}_{\mathcal{P}\left(g^{cl}_{\vartheta}\left({\boldsymbol{h}}\right) | y\right)}\delta\left( g^{cl}_{\vartheta}\left({\boldsymbol{h}}\right), \mathbb{E}_{\mathcal{P}\left(g^{cl}_{\vartheta}\left({\boldsymbol{h}}\right) | y\right)} \right)\right] \\
        - \mathbb{E}_{\mathcal{P}\left( \tilde{\boldsymbol{h}} , y\right)}\log\frac{ \tilde{\boldsymbol{h}}  \times \mu_{\boldsymbol{y}}}{|| \tilde{\boldsymbol{h}} || \times ||\mu_{\boldsymbol{y}}||} &\leq - \mathbb{E}_{\mathcal{P}\left( {\boldsymbol{h}} , y\right)}\log\frac{ {\boldsymbol{h}}  \times \mu_{\boldsymbol{y}}}{|| {\boldsymbol{h}} || \times ||\mu_{\boldsymbol{y}}||} \\
        \mathbb{E}_{\mathcal{P}\left( \tilde{\boldsymbol{h}} , y\right)}\log\frac{ \tilde{\boldsymbol{h}}  \times \mu_{\boldsymbol{y}}}{|| \tilde{\boldsymbol{h}} || \times ||\mu_{\boldsymbol{y}}||} &\geq \mathbb{E}_{\mathcal{P}\left( {\boldsymbol{h}} , y\right)}\log\frac{ {\boldsymbol{h}}  \times \mu_{\boldsymbol{y}}}{|| {\boldsymbol{h}} || \times ||\mu_{\boldsymbol{y}}||}.
    \end{aligned}
\end{equation}
As shown in the subfigure (a) of Figure \ref{fig:proofexample}, all of $\lfloor \boldsymbol{h} \rceil^k$, $\lfloor \tilde{\boldsymbol{h}} \rceil^k$, $\lfloor \boldsymbol{h} \rceil^{k+1}$, and $\lfloor \tilde{\boldsymbol{h}} \rceil^{k+1}$ denote a specific dimension of the corresponding representations. Then, we reconstruct the complete representations from the dimensions, which is achieved in the subfigure (b) of Figure \ref{fig:proofexample}. The combinations of (2, 3) and (1, 4) are hypothetical and fake, the combination (1, 3) denotes the original and unmasked representation $\boldsymbol{h}$, and (2, 4) denotes the masked representation $\tilde{\boldsymbol{h}}$. We observe that although the specific value of the vector of each dimension has a thin effect on the corresponding cosine error risk, the combined representation vector is largely sensitive to such values. As demonstrated in (1, 3) and (2, 4) in the subfigure (b), the cosine error risk of $\tilde{\boldsymbol{h}}$ and $\mu_{\boldsymbol{y}}$ is apparently less than $\boldsymbol{h}$ and $\mu_{\boldsymbol{y}}$, which supports that, given a target $\boldsymbol{y}$, the masked representation can indeed derive smaller conditional variance than the unmasked representation. The reason behind such a phenomenon is that, following Theorem \ref{the:connect}, the self-paced dimensional mask jointly enhances the gradient effect of the dimensions containing discriminative information and reduces that of the dimensions containing the confounder by adjusting the weights of different dimensions with respect to their gradient contribution to the optimization of a specific self-supervised task, e.g., contrastive learning, during training. Therefore, the dimensions containing discriminative information have relatively large weights, while the dimensions containing confounder information are partially \textit{masked}.

\subsubsection{Proof for the Tighter Bounds}
Being aware of proofs in Section \ref{prf:equal} and Section \ref{prf:inequal}, we confirm the validation of Theorem \ref{the:tighter}. Then, we bring Theorem \ref{the:tighter} into Theorem \ref{the:connect} to derive the comparison of the lower bounds of supervised cross-entropy loss that are separately obtained by the masked representation $\tilde{\boldsymbol{h}}$ and the unmasked representation $\boldsymbol{h}$:
\begin{equation} \label{eq:tighterprooflower}
	\begin{aligned}
		&\mathcal{L}_{contrast}\left(g^{cl}_{\vartheta}\left(\tilde{\boldsymbol{h}}\right)\right) - \sqrt{\Phi\left(g^{cl}_{\vartheta}\left(\tilde{\boldsymbol{h}}\right) \Big| \boldsymbol{y}\right)} - \frac{1}{2} \cdot \sum_{k \in \llbracket 1, D \rrbracket} {\Phi\left(\lfloor g^{cl}_{\vartheta}\left(\tilde{\boldsymbol{h}}\right)\rceil^k \Big| \boldsymbol{y}\right)} - \mathcal{O} \left( M^{- \frac{1}{2}} \right) \\ &\geq \mathcal{L}_{contrast}\left(g^{cl}_{\vartheta}\left(\boldsymbol{h}\right)\right) - \sqrt{\Phi\left(g^{cl}_{\vartheta}\left(\boldsymbol{h}\right) \Big| \boldsymbol{y}\right)} - \frac{1}{2} \cdot \sum_{k \in \llbracket 1, D \rrbracket} {\Phi\left(\lfloor g^{cl}_{\vartheta}\left(\boldsymbol{h}\right)\rceil^k \Big| \boldsymbol{y}\right)} - \mathcal{O} \left( M^{- \frac{1}{2}} \right).
	\end{aligned}
\end{equation}
Therefore, the lower bound obtained by the masked representation, i.e., MetaMask, is larger than the unmasked representation, i.e., typical contrastive learning methods. We further analyze the upper bound of supervised cross-entropy loss on downstream tasks and derive
\begin{equation} \label{eq:tighterproofupper}
	\begin{aligned}
		&\mathcal{L}_{contrast}\left(g^{cl}_{\vartheta}\left(\tilde{\boldsymbol{h}}\right)\right) + \sqrt{\Phi\left(g^{cl}_{\vartheta}\left(\tilde{\boldsymbol{h}}\right) \Big| \boldsymbol{y}\right)} + \mathcal{O} \left( M^{- \frac{1}{2}} \right) \\ &\leq \mathcal{L}_{contrast}\left(g^{cl}_{\vartheta}\left(\boldsymbol{h}\right)\right) + \sqrt{\Phi\left(g^{cl}_{\vartheta}\left(\boldsymbol{h}\right) \Big| \boldsymbol{y}\right)} + \mathcal{O} \left( M^{- \frac{1}{2}} \right),
	\end{aligned}
\end{equation}
where the upper bound obtained by MetaMask is smaller than the typical contrastive learning methods. Concretely, we conclude that our approach can better bound the downstream classification risk, i.e., the upper and lower bounds of supervised cross-entropy loss obtained by MetaMask are tighter than typical contrastive learning methods.

\subsection{Discussion on the Difference between Dimensional Confounder and Curse of Dimensionality} \label{app:diffcurse}
The restricted definition of the curse of dimensionality is that given a fixed number of examples, as the dimensionality of the representation increases during training, the performance of the model on downstream tasks instead degenerates, e.g., the \textit{Hughes phenomenon}. Recent researchers state a generalized definition of the curse of dimensionality from the perspective of over-fitting: given a fixed number of examples, the performance of the model degrades as the amount of \textit{computation} in the network increases, e.g., the increase of representations' dimensionality or the increase of the layers of a deep network. Concretely, the curse of dimensionality is a theory to explain a specific phenomenon of over-fitting from the perspective of sample size.

However, our \textit{dimensional confounder} is defined as a negative factor that may lead to model degradation, which is proposed from the dimensional perspective. The motivating example in Figure \ref{fig:motivmask} demonstrates that even if the deep network architecture and the dimension of the representation are consistent, the dimensional confounder still causes the performance of the model to degenerate. This experimental analysis proves that the proposed dimensional confounder and the curse of dimensionality are two different terminologies. Further, to conduct the experiments in Figure \ref{fig:motivdimincrea}, we fix the dimensionality of the representation and only increase the dimensionality of the features generated by the projection head, which is under the restriction of the redundancy-reduction technique \cite{2021Barlow}. The intuition behind such a behavior is that as the dimensions of the projected features increase, the information entropy contained in each dimension of the representation becomes more decoupled, and the total amount of task-relevant information entropy is limited, such that the dimensional confounder (i.e., the dimensions only containing task-irrelevant information) is gradually increasing. The experimental results demonstrate that as the dimensional confounder in the learned representation increases, the performance of the model is weakened. Such a phenomenon is consistent with the generalized definition of the curse of dimensionality, since the dimensionality of the representation is constant while the parameter (computation) of the deep network is increasing. Therefore, we conclude that the dimensional confounder can lead to the curse of dimensionality, and compared with the curse of dimensionality, the proposed dimensional confounder is a generalized issue. The comparisons in Table \ref{tab:convfc} and Table \ref{tab:resnet18} prove that our MetaMask can alleviate the dimensional confounder issue, and the experiments in Figure \ref{fig:robust} further support the superiority of MetaMask over current self-supervised methods to alleviate the curse of dimensionality issue.

\begin{figure}
	\centering
	\includegraphics[width=0.7\textwidth]{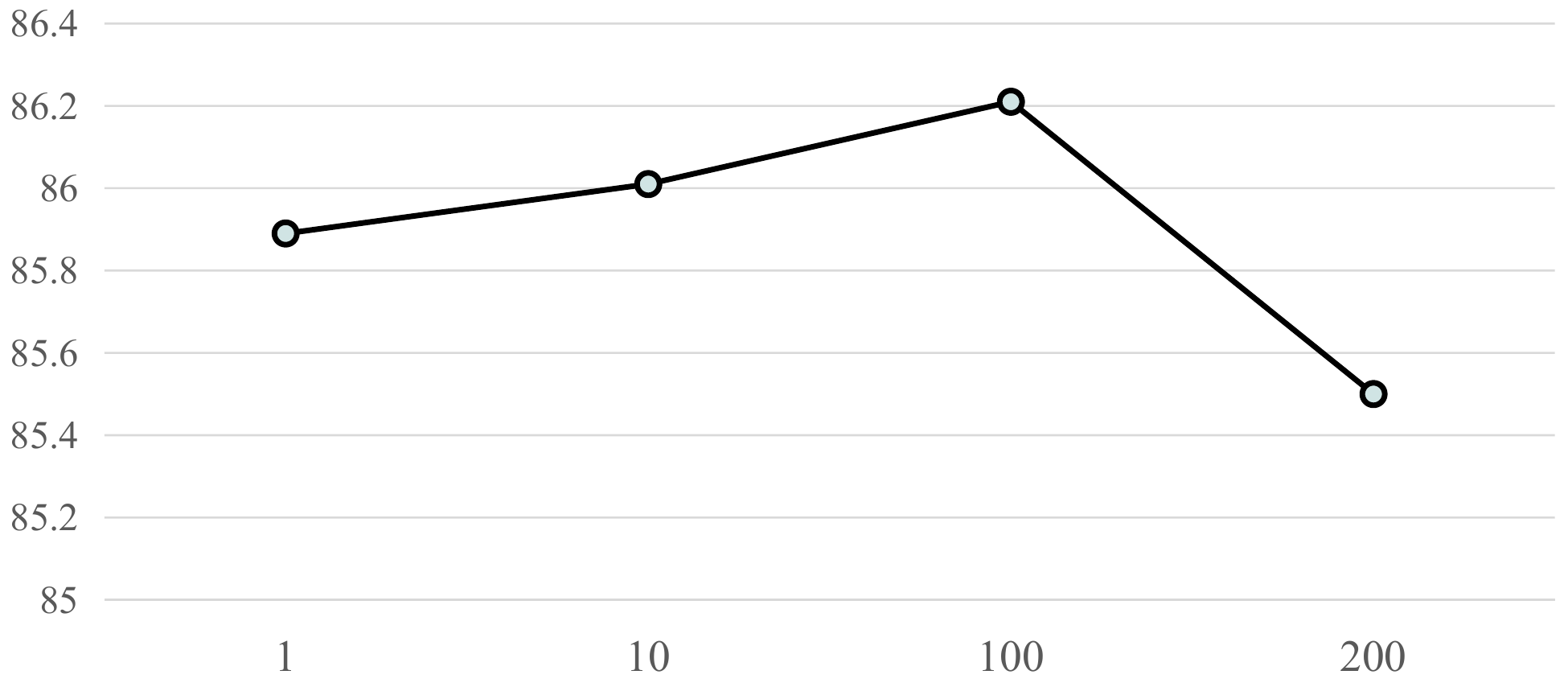}
	\vskip -0.1in
	\caption{Comparisons of MetaMask using different settings of the hyper-parameter $\alpha$.}
	\label{fig:alphastudy}
	\vspace{-0.05cm}
\end{figure}

\begin{figure}
	\centering
	\includegraphics[width=0.7\textwidth]{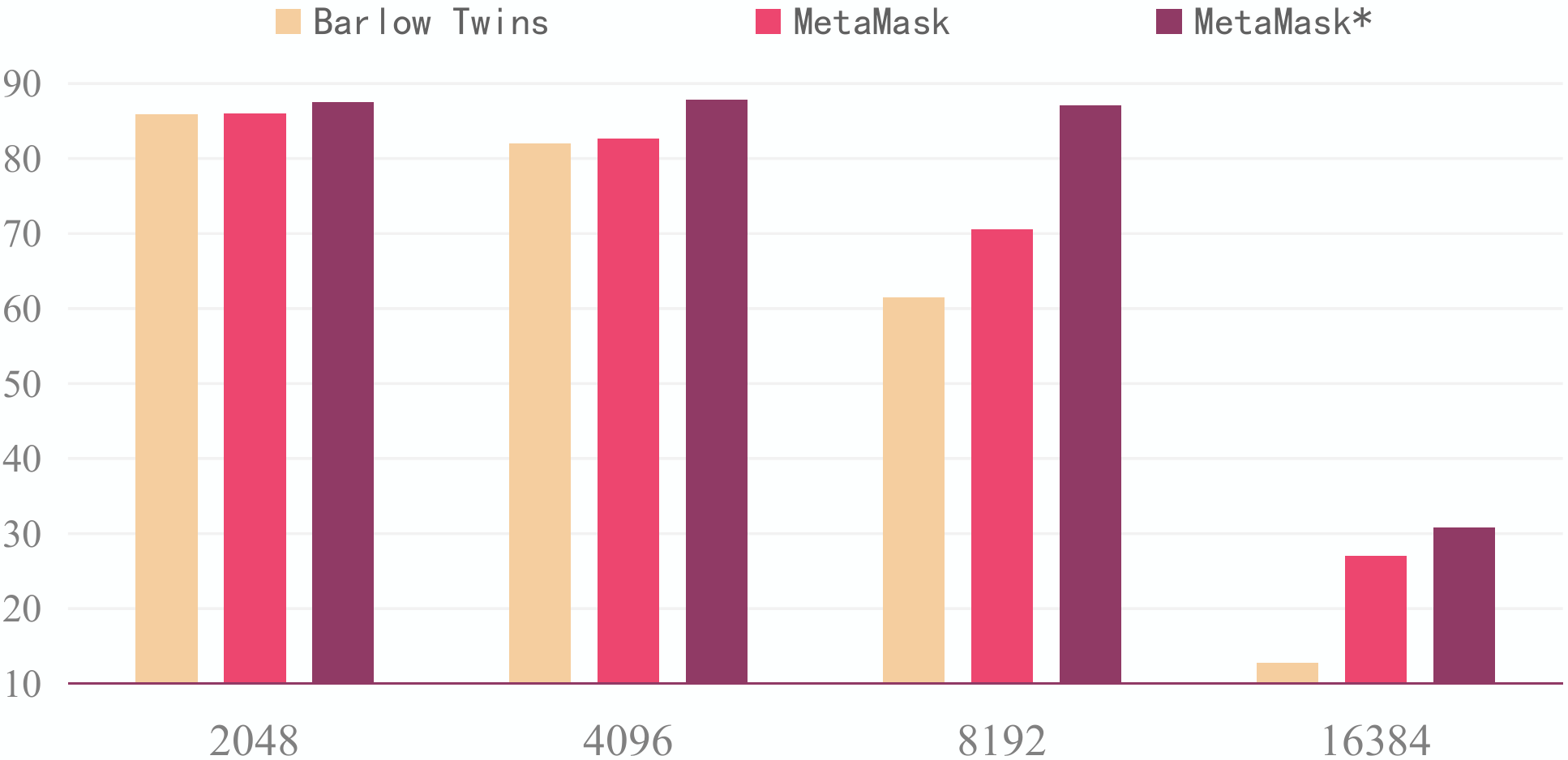}
	\vskip -0.1in
	\caption{Further comparisons of varying the projection head's dimensionality from the range of $\left\{ 2048, 4096, 8192, 16384 \right\}$, which are conducted on CIFAR-10 by using ResNet-18. $\ast$ denotes MetaMask using a trick of fixed learning rate instead of the cosine annealing strategy.}
	\label{fig:furtherrobust}
	\vspace{-0.5cm}
\end{figure}

\begin{figure}[t]
	\centering
	\includegraphics[width=0.7\textwidth]{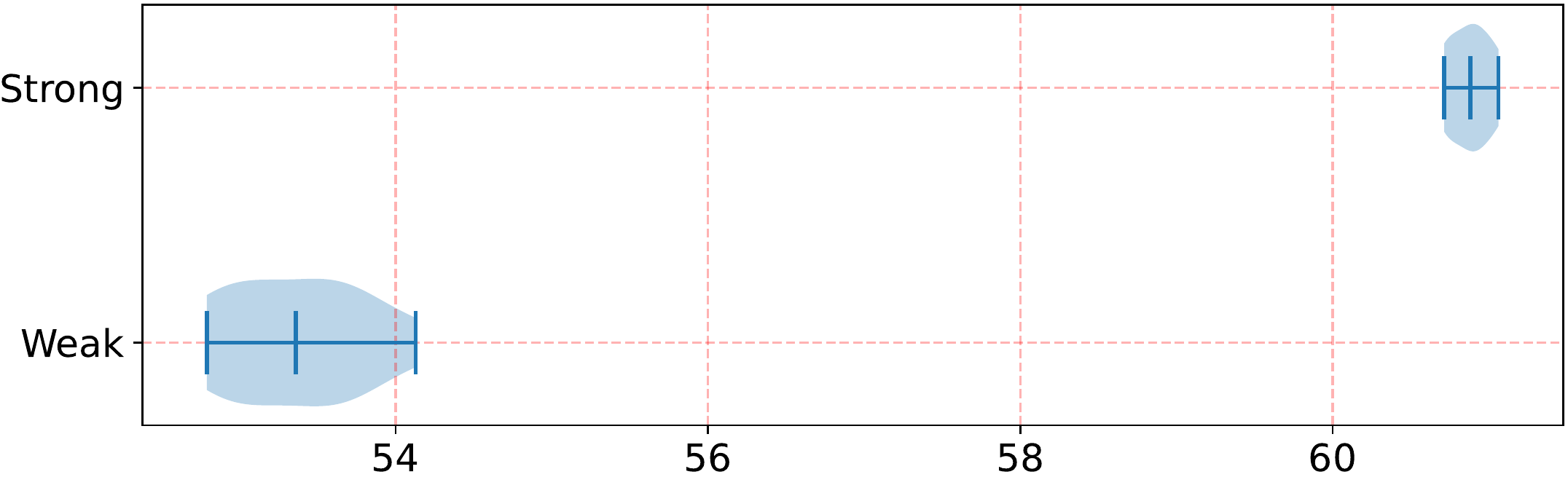}
	\vskip -0.1in
	\caption{Comparisons of BYOL using different backbones on CIFAR100 with 5\% random dimensional mask. ``Weak'' denotes the weak encoder \textit{conv}, and ``Strong'' denotes the strong encoder \textit{ResNet-18}.}
	\label{fig:weakstrong}
	\vspace{-0.05cm}
\end{figure}

\subsection{Extended Experiments} \label{app:extendedexp}
We conduct several experiments to study the intrinsic property of the proposed MetaMask.

\subsubsection{The Impact of the Hyper-Parameter $\alpha$}
As demonstrated in Figure \ref{fig:alphastudy}, we conduct comparisons of MetaMask using ResNet-18 on CIFAR-10. We follow the experimental principle in Section \ref{sec:experiments} and use KNN prediction as the evaluation approach. The comparison results show that an elaborate assignment of $\alpha$ can indeed improve the performance of MetaMask, and when $\alpha = 100$, the accuracy of MetaMask achieves the peak value. The reasons behind this curve are as follows: 1) $\mathcal{L}_{drr}$ is naturally excessively larger than $\mathcal{L}_{contrast}$ so that when $\alpha$ is not large enough, $\mathcal{L}_{contrast}$'s impact with respect to the gradient is diminished; 2) when $\alpha$ is over-large, the impact of $\mathcal{L}_{drr}$ is weakened, and the dimensional redundancy issue cannot be sufficiently addressed so that the representation may be collapsed to a trivial constant.

\subsubsection{Further Exploration of the Robustness of MetaMask}
The experimental analysis demonstrated in Section \ref{sec:valrobust} proves that MetaMask can indeed alleviate the negative impact caused by the dimensional confounder. However, MetaMask$^\ast$ achieves better performance than compared methods, including vanilla MetaMask. We provide a further exploration shown in Figure \ref{fig:furtherrobust}, and we observe that MetaMask$^\ast$ collapses abruptly at larger dimensional settings, e.g., 16384, which demonstrates our explanation of such a phenomenon in Section \ref{sec:valrobust}, i.e., the fixed learning rate trick can temporarily improve the performance of MetaMask, but this improvement is inconsistent. The results in Figure \ref{fig:furtherrobust} further show that with or without this trick, the proposed MetaMask is robust against the dimensional confounder.

\subsubsection{Impacts of Encoder's Ability towards MetaMask} \label{app:encoderimpact}
From the experimental results reported in Section \ref{sec:experiments}, we observe that MetaMask improves the benchmark methods with weak encoders by excessively significant margins, but the improvement with strong encoders is relatively limited. Our consideration behind this phenomenon is that although representations learned by both weak and strong encoders (without our method) may contain dimensional confounders, the strong encoder can better capture semantic information so that the dimensional confounder of the learned representation is naturally less, and the useful discriminative information is much more than the representation learned by the weak encoder. \cite{wang2022chaos} provide the theorem and corresponding proof to demonstrate that the contrastive loss can bound the cross-entropy loss in downstream tasks. Strong encoders better minimize the contrastive loss, and thus the representations learned by strong encoders contain more semantic information, and accordingly, fewer dimensional confounders. However, from the results reported in Table \ref{tab:resnet18}, our method can still improve the benchmark methods.

We further conduct experiments by imposing random dimensional masks on the learned representations for weak and strong encoders. The results are reported in Figure \ref{fig:weakstrong}. The comparisons demonstrate that under the consistent setting of 5\% random dimensional mask rate, the results of the weak encoder (conv) range from 52.79 to 54.13, and the results of the strong encoder (ResNet-18) range from 60.71 to 61.06. Note that we conduct 10 trials for each experiment to achieve unbiased results. The quality of the strong encoder's representation is apparently better, and the performance of the strong encoder is more stable (the variance of the strong encoder's results is smaller), which proves that the representation learned by the strong encoder contains fewer dimensional confounders, and each dimension contains more discriminative information. Therefore, our method improves weak encoders more than strong encoders.

\begin{table}[t]
	\tiny
	\renewcommand\arraystretch{1.1}
	\vskip 0.1in
	\caption{The complexity comparisons between MetaMask and benchmark methods on CIFAR10. Note that for fair comparisons, this experiment is based on 1 GPU of NVIDIA Tesla V100.}
	\vskip -0.14in
	\label{tab:complexity}
	\setlength{\tabcolsep}{9.5pt}
	\begin{center}
		\begin{small}
			\begin{tabular}{lcc}
				\toprule
				Methods & Parameters & Training time cost for an epoch \\
				\midrule
				ResNet-18 & 11.2M & - \\
                SimCLR & 13M & 70s \\
                Barlow Twins & 22.7M & 80s \\
                SimCLR + Barlow Twins & 24.6M & 85s \\ \rowcolor{mygray}
                MetaMask & 24.6M + 512 & 210s \\
				\bottomrule
			\end{tabular}
		\end{small}
	\end{center}
	\vskip -0.2in
\end{table}

\begin{table}[t]
	\tiny
	\renewcommand\arraystretch{1.1}
	\vskip 0.1in
	\caption{The comparisons between MetaMask and benchmark methods on the CIFAR10 dataset by using the same total time costs. Note that for fair comparisons, this experiment is based on 1 GPU of NVIDIA Tesla V100.}
	\vskip -0.14in
	\label{tab:samecost}
	\setlength{\tabcolsep}{9.5pt}
	\begin{center}
		\begin{small}
			\begin{tabular}{lccc}
				\toprule
				Methods & Epoch & Training time cost & Accuracy \\
				\midrule
				SimCLR & 2400 & 46h & 81.75 \\
                Barlow Twins & 2100 & 46h & 85.71 \\
                SimCLR + Barlow Twins & 2000 & 47h & 85.79 \\ \rowcolor{mygray}
                MetaMask & 800 & 46h & 86.01 \\
				\bottomrule
			\end{tabular}
		\end{small}
	\end{center}
	\vskip -0.2in
\end{table}

\subsubsection{Training Complexity of MetaMask}
To compare the training complexity of MetaMask and benchmark methods, we conduct experiments on CIFAR10 by using ResNet-18 as the backbone network. The results are demonstrated in Table \ref{tab:complexity}, which shows that the parameter number used by MetaMask is close to the ablation model, i.e., SimCLR + Barlow Twins, and Barlow Twins. Compared with Barlow Twins, SimCLR + Barlow Twins only adds a MLP with several layers as the projection head for the contrasting learning of SimCLR. For MetaMask, we only add a learnable dimensional mask $\mathcal{M}$ in the network of MetaMask. Note that to decrease the time complexity of MetaMask, we create an additional parameter space to save the temporary parameters for computing second-derivatives, but such cloned parameters are not included in the computation of the network during training.

For the time complexity, due to the learning paradigm of meta-learning (second-derivatives technique), MetaMask has larger time complexity than benchmark methods (including the ablation model SimCLR + Barlow Twins). We further conduct experiments to evaluate the performance of the compared methods using similar total training time costs. The results are reported in Table \ref{tab:samecost}, which demonstrates that MetaMask still achieves the best performance, and MetaMask can also beat the ablation model SimCLR + Barlow Twins. This proves that although the time complexity of MetaMask is a little bit high, the improvement of MetaMask is consistent and solid. We also include these results and the corresponding analysis in Appendix of the rebuttal revised version.

\begin{table}[t]
	\tiny
	\renewcommand\arraystretch{1.1}
	\vskip 0.1in
	\caption{The comparisons of Barlow Twins using different dropout ratios on CIFAR10 with ResNet-18.}
	\vskip -0.14in
	\label{tab:dropout}
	\setlength{\tabcolsep}{5.5pt}
	\begin{center}
		\begin{small}
			\begin{tabular}{ccc}
				\toprule
				\multirow{2}*{Dropout ratio} & Dropout shared & \multirow{2}*{Accuracy} \\
				& between views & \\
				\midrule
				0 & No & 85.72 \\
                0.01 & No & 85.76 \\
                0.05 & No & 86.11 \\
                0.1 & No & 85.72 \\
                0.2 & No & 85.13 \\
                0.5 & No & 81.54 \\
                0.05 & Yes & 31.17 \\
                0.1 & Yes & 37.15 \\
                0.5 & Yes & 31.62 \\
				\bottomrule
			\end{tabular}
		\end{small}
	\end{center}
	\vskip -0.2in
\end{table}

\subsubsection{Exploration of Barlow Twins using the Dropout Trick}
We apply dropout to randomly set the features generated by the backbone network to 0 with a given probability. We use two different settings: 1) features from different views are randomly set to 0 independently with 6 probabilities, including 0.01, 0.05, 0.1, 0.2, and 0.5, which is shown as the setting of dropout shared between views is ``No''; 2) for the same sample, we set the same channels of the features from different views to 0. In detail, we let the features from the first view pass the dropout layer, and then record the channels which are set to 0. Finally, we set the same feature channels of the second view to 0 and multiply the features with 1/(1-p), where ``p'' refers to the probability of dropout. In this case, we use three probabilities: 0.05, 0.1, and 0.5. This setting is shown as ``Yes'' for the dropout shared between views. The results are reported in Table \ref{tab:dropout}. We observe that models trained by following the second setting are collapsed, and these models far underperform MetaMask. For the first setting, the performance trend is in accordance with our expectation, using several well-chosen dropout ratios can improve the performance of Barlow Twins to a certain extent, but the improvement is inconsistent. Furthermore, even the best model of Barlow Twins using the dropout trick cannot achieve comparable performance to MetaMask, since as shown in Table \ref{tab:resnet18}, MetaMask achieves 87.53 on CIFAR10 with ResNet-18.

\begin{table}[t]
	\tiny
	\renewcommand\arraystretch{1.1}
	\vskip 0.1in
	\caption{Top-1 and top-5 accuracies (\%) under linear evaluation on ImageNet with ResNet-50 encoder. MetaMask is performed based on Barlow Twins and SimCLR. Top-3 best methods are underlined.}
	\vskip -0.14in
	\label{tab:inresnet50}
	\setlength{\tabcolsep}{5.5pt}
	\begin{center}
		\begin{small}
			\begin{tabular}{lcc}
				\toprule
				\multirow{2}*{Method} & \multicolumn{2}{c}{Accuracy} \\
				\cline{2-3}
				& Top-1 & Top-5 \\
				\midrule
                Supervised & \underline{76.5} & - \\
                \midrule
                MoCo & 60.6 & - \\
                SimCLR & 69.3 & 89.0 \\
                SimSiam & 71.3 & - \\
                BYOL & \underline{74.3} & \underline{91.6} \\
                Barlow Twins & 73.2 & \underline{91.0} \\ \rowcolor{mygray}
                MetaMask & \underline{73.9} & \underline{91.4} \\
				\bottomrule
			\end{tabular}
		\end{small}
	\end{center}
	\vskip 0.in
\end{table}

\subsubsection{Performing MetaMask on ImageNet with ResNet-50}
Training one trial of MetaMask on ImageNet is excessively time-consuming for the adopted server. It is hard to impose sufficient hyperparameter tuning experiments, because tuning parameters on the validation set and then retraining is time-consuming. For the current version, we principally adopt the hyperparameter of MetaMask on ImageNet-200 with ResNet-18 for the experiments on ImageNet with ResNet-50. We report the results on Table \ref{tab:inresnet50}, which demonstrates that MetaMask achieves the top-3 best performance. Note that the major results are reported by \cite{2021Barlow}, and the implementation of MetaMask is based on SimCLR and Barlow Twins. For comparisons with the main baselines, MetaMask beats SimCLR by 4.6 on top-1 accuracy and 2.4 on top-5 accuracy, and MetaMask beats Barlow Twins by 0.7 on top-1 accuracy and 0.4 on top-5 accuracy. The improvement of MetaMask is in accordance with our observation in Appendix \ref{app:encoderimpact}. Furthermore, ImageNet-200 is a truncated dataset of ImageNet, so the domain shift between these datasets is slight. The comparisons in Table \ref{tab:convfc}, and Table \ref{tab:resnet18} demonstrate that the proposed MetaMask can still improve the benchmark methods in large-scale datasets.

\begin{figure}[t]
	\centering
	\includegraphics[width=0.7\textwidth]{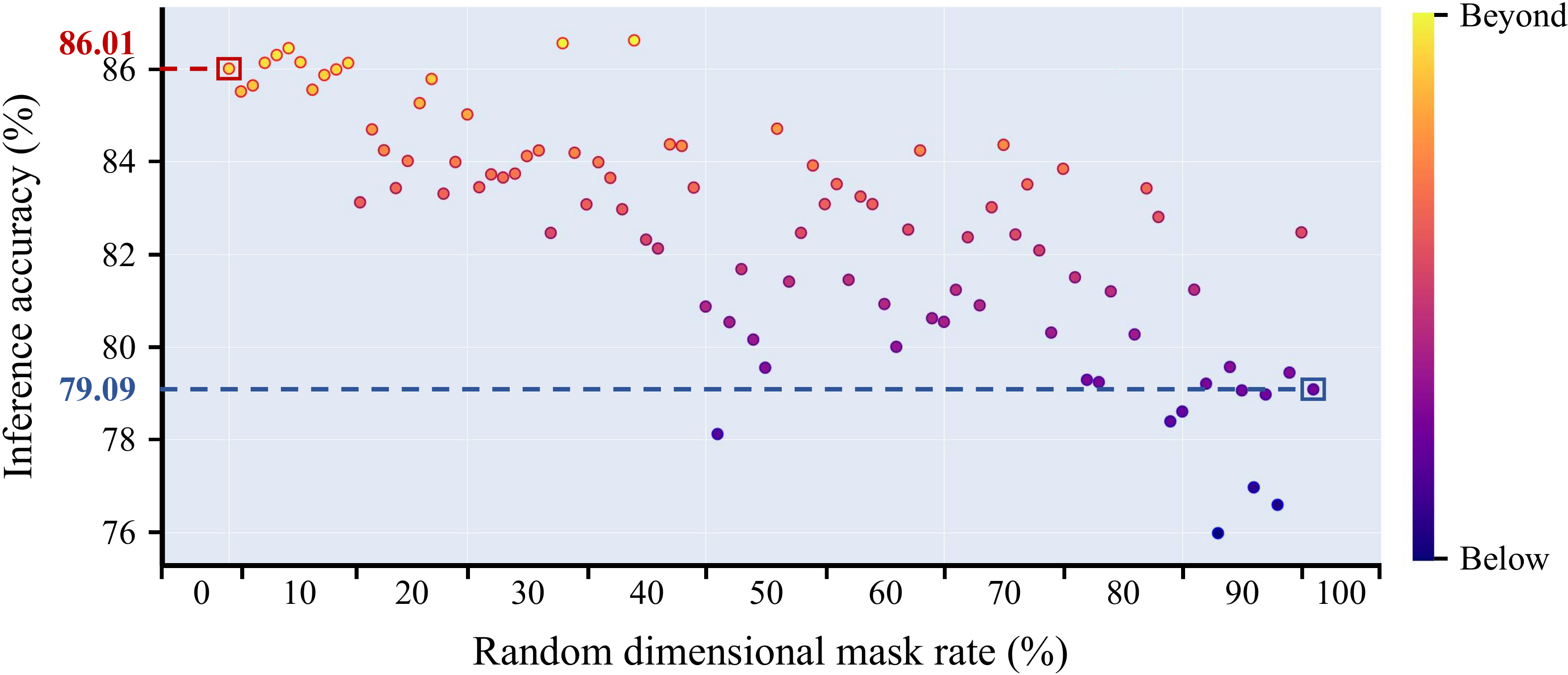}
	\vskip -0.1in
	\caption{Experimental scatter diagrams obtained by MetaMask with randomly masked dimensions on CIFAR-10 with ResNet-18. We collect the final dimensional weight matrix $\mathcal{M}$ and then choose dimensions with weights below average as the masked dimensions. These dimensions are considered to be associated with the dimensional confounders, and we impose random dimensional masks on these dimensions. For fair comparisons, 10 trials are conducted per mask rate, except for 0\% and 100\% mask rates, and every single point denotes a trial. The trial boxed by \textit{\textcolor[RGB]{205,71,119}{red} lines} presents the performance achieved by the unmasked representation of MetaMask, and the trial boxed by \textit{\textcolor[RGB]{47,85,161}{blue} lines} presents the performance achieved by MetaMask's representation where all dimensions containing confounders are masked.}
	\label{fig:ablmask}
	\vspace{-0.05cm}
\end{figure}

\subsubsection{Exploration of Confounders Contained by Masked Dimensions} \label{app:confounderdeep}
MetaMask trains $\mathcal{M}$ by adopting a meta-learning-based training approach, which ensures that $\mathcal{M}$ can partially mask the ``gradient contributions'' of dimensions containing task-irrelevant information and further promote the encoder to focus on learning task-relevant information. So, MetaMask only performs the gradient mask during training instead physically masking dimensions in the test. We provide theoretical explanation and proofs in Appendix \ref{app:discusion} and Appendix \ref{app:proof}. The reason behind our behavior is that even dimensions that contain dimensional confounders are also possible to contain discriminative information so that lowering the gradient contribution of such dimensions can not only prevent the over-interference of the dimensional confounders on the representation learning but also preserve the acquisition of the information of these dimensions. Accordingly, the foundational idea behind self-supervised learning is to learn ``general'' representation that can be generalized to various tasks. MetaMask introduces a meta-learning-based approach to train the dimensional mask $\mathcal{M}$ with respect to improving the performance of contrastive learning. However, the theorems, proposed by \cite{wang2022chaos}, only prove that the contrastive learning objective is associated with the downstream classification task, while there is no theoretical evidence to demonstrate the connection between contrastive learning objective and other downstream tasks. Therefore, we consider not directly masking the dimensions containing dimensional confounders in the test.

Furthermore, we conduct experiments to explore the performance of the variant that directly masks these dimensions in the test, which is demonstrated in Figure \ref{fig:ablmask}. For the exploration of our masking scheme and its variants, we conduct experiments as follows: after training, we collect the final dimensional weight matrix $\mathcal{M}$ and then choose dimensions with weights below average as the masked dimensions. These dimensions are considered to be associated with dimensional confounders. To prove whether these dimensions have confounders, we perform random dimensional masking to these dimensions, and when the masking rate is 100\%, the model turns to the variant that directly masks all these dimensions in the test. The experiments are based on SimCLR + MetaMask. Note that we conduct 10 trials per mask rate (except for 0\% and 100\% mask rates) for fair comparisons. We observe that the original MetaMask (i.e., mask rate is 0\%) achieves the best performance on average, and MetaMask outperforms the variant masking the dimensions with confounders by a significant margin, which proves that our proposed approach, i.e., masking the ``gradient contributions'' of dimensions in the training is more effective than the compared approach, i.e., directly masking dimensions in the test. While, several trials with specific mask rates demonstrate better performance than MetaMask, which proves that the dimensions filtered by MetaMask indeed contain dimensional confounders. Additionally, observing the results reported in Figure \ref{fig:motivmask} and the results in Figure \ref{fig:ablmask}, we find that the results achieved by the proposed variants are better than Barlow Twins with random dimensional masks on average, which can further prove the filtered dimensions containing confounders and that MetaMask indeed assigns lower gradient weights to the dimensions containing confounders.

Additionally, we conduct experiments to directly impose a linear probe on the whole masked or unmasked features. Note that the definition of masked features is mentioned above, and the experiments are conducted on CIFAR10 with ResNet-18. As shown in Figure \ref{fig:ablmask}, a linear probe is solely imposed on the whole ``unmasked'' features (without ``masked'' features), which is the same as the 100\% mask rate variant for the ``masked’’ features above, and the result is 79.09. We provide the corresponding reasons and analyses above. Additionally, such results may be due to the masking scheme for these experiments, i.e., collecting the final dimensional weight matrix $\mathcal{M}$ and then masking the dimensions with weights below average, which is dramatically different from MetaMask's behavior, and this scheme is only used for this exploration. The result achieved by a linear probe on the whole ``masked'' features is 56.30, which demonstrates our statement that dimensions containing dimensional confounders are also possible to contain discriminative information, because the achieved accuracy is not under 10. The result also proves that MetaMask can indeed assign lower gradient weights to the dimensions containing confounders, since such a model far underperforms MetaMask (86.01).

\begin{figure}[t]
	\centering
	\includegraphics[width=0.7\textwidth]{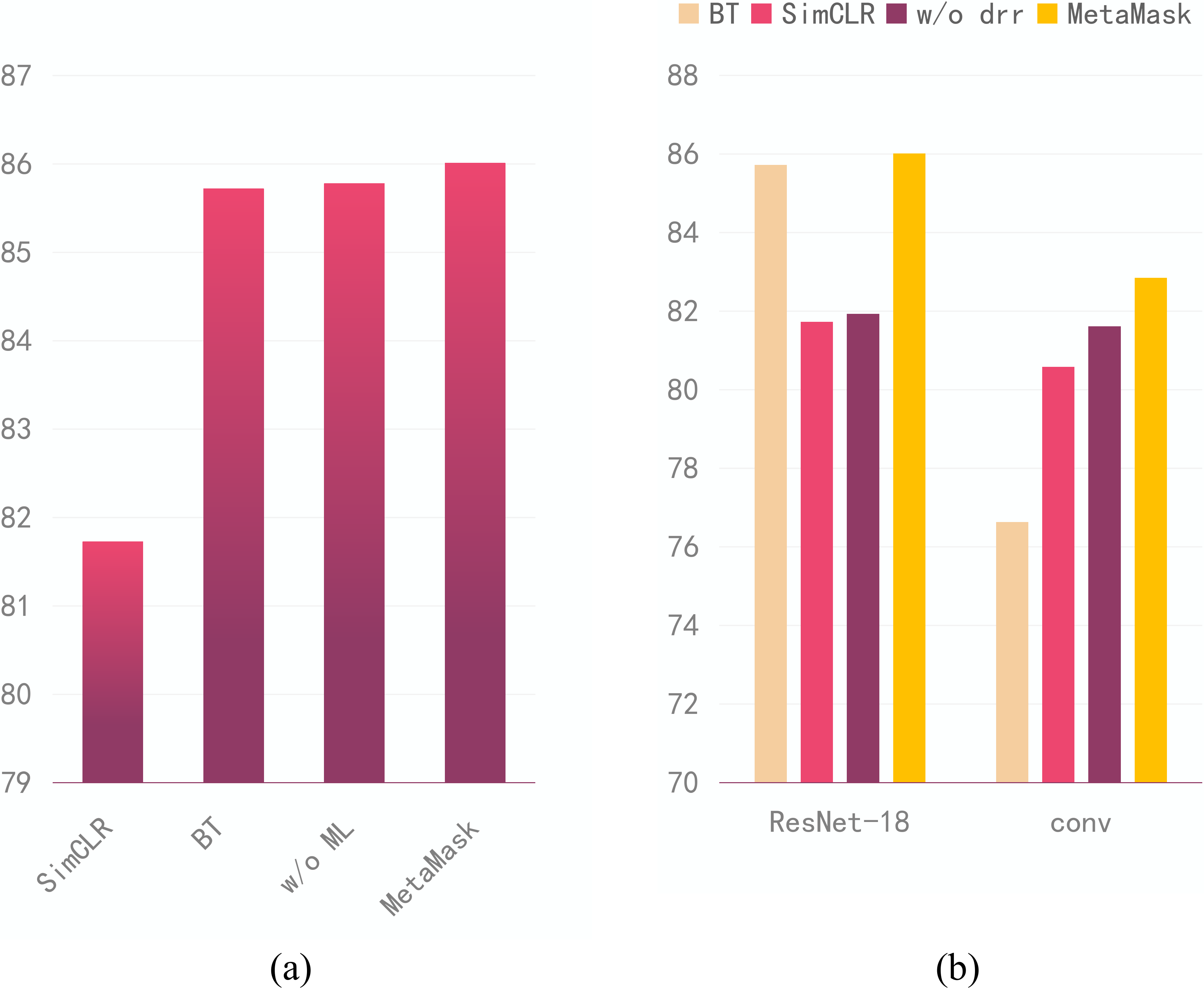}
	\vskip -0.1in
	\caption{Ablation studies obtained by MetaMask on CIFAR-10 with ResNet-18 and conv encoders, where ``BT'' denotes Barlow Twins, ``w/o ML'' denotes the ablation model without the proposed $\mathcal{M}$ and the corresponding meta-learning-based learning paradigm, i.e., SimCLR + Barlow Twins, and ``w/o drr'' denotes the ablation model without the dimensional redundancy reduction loss $\mathcal{L}_{drr}$. (a) We conduct experiments to prove the effectiveness of the proposed dimensional mask $\mathcal{M}$ on CIFAR10 with ResNet-18, and the results show that only w/o ML (SimCLR + Barlow Twins) can not improve the performance of the model by a significant margin, and the proposed approach is crucial to the performance improvement. (b) We conduct experiments on CIFAR10 with ResNet-18 and conv, respectively. The results further prove the effectiveness of the proposed $\mathcal{M}$ and $\mathcal{L}_{drr}$.}
	\label{fig:abltest}
	\vspace{-0.1cm}
\end{figure}

\subsubsection{Further Ablation Study of MetaMask}
Barlow Twins handles dimensional redundancy but suffers dimensional confounders. MetaMask mitigates dimensional confounders by learning and applying a dimensional mask. We conduct experiments to demonstrate our statement, and the results are reported in Figure \ref{fig:abltest}. For the experiments shown in Figure \ref{fig:abltest} (a), we demonstrate the effectiveness of the proposed dimensional mask $\mathcal{M}$ and the corresponding meta-learning-based training paradigm by directly removing such approaches. The results show that the sole w/o ML only improves Barlow Twins by 0.04, while MetaMask can improve BT by 0.29, which proves the proposed $\mathcal{M}$ is pivotal to the performance promotion. In Figure \ref{fig:abltest} (b), to verify whether there would be performance gain only from alleviating dimensional confounders without $\mathcal{L}_{drr}$, we evaluate the performance of w/o drr and MetaMask. We observe that for the experiments with ResNet-18, w/o drr (without Barlow Twins) improves SimCLR by 0.2 but cannot reach the performance of BT, and MetaMask can improve both SimCLR and BT by 4.28 and 0.29, respectively. For the experiments with conv, w/o drr improves SimCLR by 1.03 and also outperforms BT by 4.08, and MetaMask improves both SimCLR and BT by 2.27 and 6.22, respectively. Concretely, the performance of w/o drr is related to the performance of SimCLR, and it can always improve SimCLR but underperform the complete MetaMask. MetaMask has consistent best performance by using different encoders.

We consider that $\mathcal{L}_{drr}$ (Barlow Twins loss) could exacerbate dimensional confounders, but as our discussion in Appendix \ref{app:confounderdeep}, i.e., dimensions containing confounders are also possible to contain discriminative information, more dimensions with confounders due to $\mathcal{L}_{drr}$ may also carry more discriminative information. Likewise, the model without $\mathcal{L}_{drr}$ may generate representations with over-redundant dimensions so that the total amount of available discriminative information will decrease. However, roughly using the representations with complex dimensional information (without $\mathcal{L}_{drr}$) may result in insufficient discriminative information mining, e.g., w/o ML can only outperform SimCLR by a limited margin. Our proposed MetaMask effectively avoids the appearance of such an undesired phenomenon by leveraging $\mathcal{M}$ and the corresponding meta-learning-based training paradigm, which is supported by the empirical results.

\end{document}